\documentclass[conference]{IEEEtran}
\IEEEoverridecommandlockouts
\usepackage{cite}
\usepackage{amsmath,amssymb,amsfonts}
\usepackage{algorithmic}
\usepackage{graphicx}
\usepackage{textcomp}
\usepackage{algorithm}
\usepackage{amsmath}
\usepackage{algorithmic}
\usepackage{subfigure}
\usepackage{booktabs} 
\usepackage{array}    
\usepackage{tabularx}
\def\BibTeX{{\rm B\kern-.05em{\sc i\kern-.025em b}\kern-.08em
    T\kern-.1667em\lower.7ex\hbox{E}\kern-.125emX}}

\usepackage{multirow}
\usepackage[table,xcdraw]{xcolor}
\usepackage{hyperref}

\newcommand{\wl}[1]{{\color{black}#1}}

\begin{document}

\title{Efficient Graph Condensation via Gaussian Process\\
}

\author{\IEEEauthorblockN{Lin Wang}
\IEEEauthorblockA{\textit{The Hong Kong Polytechnic University} \\
Hong Kong SAR. \\
comp-lin.wang@connect.polyu.hk}
\and
\IEEEauthorblockN{Qing Li* \thanks{* Corresponding author.}}
\IEEEauthorblockA{\textit{The Hong Kong Polytechnic University} \\
Hong Kong SAR. \\
csqli@comp.polyu.edu.hk}
}

\maketitle

\begin{abstract}
Graph condensation reduces the size of large graphs while preserving performance, addressing the scalability challenges of Graph Neural Networks caused by computational inefficiencies on large datasets. Existing methods often rely on bi-level optimization, requiring extensive GNN training and limiting their scalability. 
\wl{To tackle these issues, we propose Graph Condensation via Gaussian Process (GCGP), a novel and efficient framework that optimizes a compact, high-fidelity condensed graph, enabling effective training of various GNNs with reduced computational cost.}
GCGP utilizes a Gaussian Process (GP), with the condensed graph serving as observations, to estimate the posterior distribution of predictions. This approach eliminates the need for the iterative and resource-intensive training typically required by GNNs.
To enhance the capability of the GCGP in capturing dependencies between function values, we derive a specialized covariance function that incorporates structural information. This covariance function broadens the receptive field of input nodes by local neighborhood aggregation, thereby facilitating the representation of intricate dependencies within the nodes.
To address the challenge of optimizing binary structural information in condensed graphs, Concrete random variables are utilized to approximate the binary adjacency matrix in a continuous counterpart. This relaxation process allows the adjacency matrix to be represented in a differentiable form, enabling the application of gradient-based optimization techniques to discrete graph structures.
Experimental results show that the proposed GCGP method efficiently condenses large-scale graph data while preserving predictive performance, addressing the scalability and efficiency challenges.
\wl{The implementation of our method is publicly available at \href{https://github.com/WANGLin0126/GCGP}{https://github.com/WANGLin0126/GCGP}}.

\end{abstract}

\section{Introduction}
Graph-structured data, consisting of nodes and edges, is a versatile representation used to model various real-world systems, including social networks\cite{aichner2021twenty, verduyn2020social}, transportation infrastructures\cite{diao2021impacts, gu2020performance}, and molecular structures\cite{wieder2020compact, reiser2022graph}. Its ability to capture relationships and interactions makes it a powerful framework for describing complex systems.
To effectively extract meaningful insights from graph-structured data, graph neural networks (GNNs)\cite{brody2021attentive, xu2018powerful, sun2021heterogeneous} have been developed as a specialized class of deep neural networks. GNNs are designed to process graph data by leveraging a message-passing mechanism\cite{wu2020comprehensive}. Through iterative aggregation of information from neighboring nodes, GNNs expand the receptive field of nodes, enabling them to represent both local and global graph structures\cite{wu2019simplifying}. This capability has made GNNs highly effective in numerous graph mining tasks\cite{chakrabarti2022graph,kang2020inform, sun2023all}.
Despite their success, the application of GNNs faces significant challenges due to the large scale of graph data\cite{hu2021ogb, duan2022comprehensive, sun2023counter}. Training and deploying GNNs often require substantial computational resources, including extensive memory, and prolonged GPU usage. 
Furthermore, the requirements of training iterations, hyperparameter tuning, and neural architecture search significantly limit their practical deployment.

\begin{figure}
    \centering
    \includegraphics[width=0.8\linewidth]{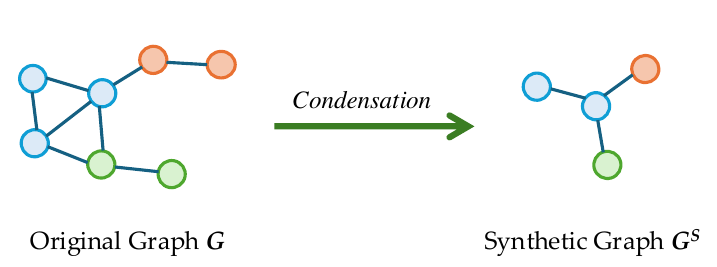}
    \caption{Graph condensation reduces a large graph dataset $G$ into a smaller one $G^{\mathcal{S}}$ with fewer nodes, while preserving essential information.}
    \label{fig:cond}
\end{figure}

Graph condensation~\cite{jin2021graph, jin2022condensing} techniques have been introduced to address the computational challenges associated with GNN training. These methods condense the original dataset into a smaller synthetic dataset while retaining critical information. The resulting condensed dataset enables more efficient GNN training by reducing both computational costs and training time. \wl{By mitigating the resource-intensive nature of GNN training, graph condensation techniques facilitate the scalability and broader applicability of GNNs~\cite{long2025efficient, gong2025scalable}}. Current graph condensation methods are commonly framed as bi-level optimization problems~\cite{jin2021graph, zhang2024navigating}, where the inner loop trains the GNN using condensed data, and the outer loop updates the condensed data.

While these approaches have achieved notable progress, they face significant challenges, particularly in computational inefficiency~\cite{wang2024fast} caused by the extensive iterative training of GNNs. The nested loops of bi-level optimization require iterative updates for both GNN parameters and condensed data, resulting in slow convergence. Moreover, to improve the generalization of the condensed data across GNNs with diverse initializations, the GNN in the inner loop often involves thousands of parameter reinitializations. This excessive computational demand undermines the efficiency of graph condensation, to the extent that training models directly on raw data may become more practical than using condensed data.


Gaussian Process (GP)~\cite{williams2006gaussian, lee2017deep}, a non-parametric method, eliminates the need for complex iterative procedures during model training. This characteristic positions GPs as a computationally efficient alternative to GNNs in graph condensation tasks.
A GP is defined by its mean function and covariance function, which together enable the posterior outputs for input data by leveraging prior knowledge and observations. Incorporating GPs into the graph condensation process eliminates the iterative training typically required by GNN models. Consequently, this approach enhances computational efficiency while simultaneously avoiding the dependency of the condensed data on parameter initialization, thereby improving the robustness and overall effectiveness of the condensation process.

Employing GPs for prediction poses significant challenges.
In graph-structured data, node features describe node attributes, while structural information encodes dependencies and interaction patterns through topological relationships. Combining these aspects challenges model design, as covariance functions must capture both feature similarity and graph context. Structural information is vital for modeling graph data, as it reflects both local and global topological characteristics. Neglecting it can hinder a model's ability to represent graph semantics, making the integration of node features and structural information a key challenge.
A second challenge stems from the discrete nature of graph structures, such as the adjacency matrix, which is non-differentiable and incompatible with gradient-based optimization methods. This limitation hinders the efficient optimization of condensed graph data.
To address these challenges, we propose a covariance function that integrates structural information from graph topology. This function supports neighborhood message passing, expanding node receptive fields and enhancing dependency modeling while minimizing computational costs~\cite{lee2017deep, wu2019simplifying}. This approach systematically improves GP predictive performance.
For optimizing structural information, we relax the binary adjacency matrix~\cite{jin2022condensing} into a differentiable form using concrete random variables~\cite{maddison2016concrete}. As the temperature approaches zero, the relaxed adjacency matrix converges to a binary form. This relaxation enables gradient-based optimization to refine the graph structure effectively.
\begin{figure}
    \centering
    \includegraphics[width=0.8\linewidth]{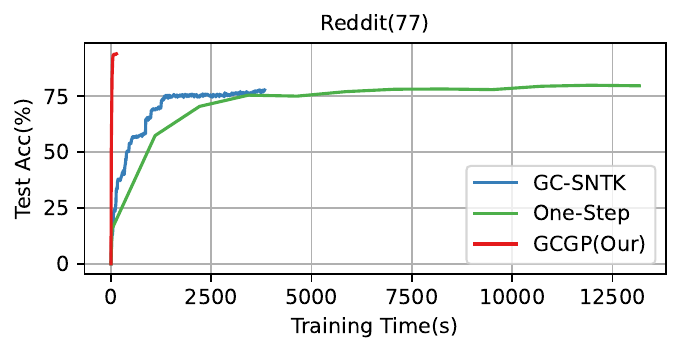}
    \caption{Condensation time and accuracy comparison on Reddit dataset. By condensing 153,431 training nodes into a graph with only 77 synthetic nodes, the GCGP achieves an accuracy of 93.9\% on the test set, which is comparable to the accuracy obtained using the full training set on GCN, while also demonstrating the fastest runtime. For more experimental results about the condensation efficiency, please refer to Section~\ref{sec:effi}.}
    \label{fig:reddit}
\end{figure}
As shown in Figure~\ref{fig:reddit}, our method condenses the training set of the Reddit~\cite{hamilton2017inductive} dataset to 77 nodes, representing only 0.05

In this paper, we propose a novel GP-based graph condensation framework for node classification tasks, termed \textbf{G}raph \textbf{C}ondensation via \textbf{G}aussian \textbf{P}rocesses (\textbf{GCGP}). The framework leverages a GP model as the predictive component, where the condensed data serve as observations. To enhance the predictive capabilities of our model, we propose a specialized covariance function that effectively captures dependencies between test inputs and observations. Furthermore, we employ concrete random variables to relax the binary adjacency matrix into a differentiable form, thereby facilitating gradient-based optimization. The discrepancy between the GP predictions and the ground truth is then leveraged to guide the optimization, particularly in optimizing the synthetic graph's structure.

The key contributions of this work are as follows:
\begin{itemize}

\item We propose a graph condensation method using GP with a condensed graph as observations to improve graph condensation efficiency.

\item We design a covariance function for graph-structured data, leveraging local neighborhood aggregation to capture structural information.

\item We introduce the binary concrete relaxation to optimize the adjacency matrix, enabling gradient-based optimization of graph structures.

\item We validate the proposed method through experiments on diverse datasets and baselines, demonstrating its efficiency and adaptability across various GNN architectures.

\end{itemize}

\section{Related Work}
\noindent \textbf{Gaussian Process}. The Gaussian Process, a widely recognized non-parametric model, has been extensively studied and serves as a foundational tool in machine learning and statistical modeling.
Williams~\cite{williams1996computing} first derived the analytical forms of the covariance matrix for single-hidden-layer infinite-width neural networks, laying the groundwork for subsequent theoretical developments. Building on this, Lee et al.\cite{lee2017deep} established a connection between infinite-width deep neural networks and Gaussian processes, providing a formal probabilistic framework. Expanding this line of research, Jacot et al.\cite{jacot2018neural} introduced the Neural Tangent Kernel, which offers insights into the training dynamics of infinitely wide networks. Furthermore, Matthews et al.~\cite{matthews2018gaussian} demonstrated the emergence of Gaussian process behavior in such networks, reinforcing the theoretical underpinnings of their behavior.

\noindent \textbf{Graph Condensation}.  
Dataset condensation~\cite{wang2018dataset,zhao2020dataset} aims to condense large-scale datasets into smaller synthetic datasets while ensuring comparable model performance. The bi-level optimization framework DC~\cite{wang2018dataset} builds on the concept of optimizing image pixels as parameters via gradients~\cite{maclaurin2015gradient}. Key evaluation criteria for DC include gradient matching~\cite{zhao2020dataset, zhao2021dataset, lee2022dataset}, feature alignment~\cite{wang2022cafe}, training trajectory matching~\cite{cazenavette2022dataset}, and distribution matching~\cite{zhao2023dataset}. Additionally, \cite{nguyen2020dataset} introduced a meta-learning approach for DC, drawing on infinite-width neural network theory~\cite{du2019graph, jacot2018neural, arora2019exact, li2019enhanced}.

Extending the concept of dataset condensation to graph data, graph condensation encompasses two primary tasks: node-level condensation and graph-level condensation~\cite{jin2021graph, jin2022condensing, gao2025graph}. Node-level condensation synthesizes a smaller graph with fewer nodes, while graph-level condensation reduces a set of graphs into a smaller, synthetic set. Jin et al.\cite{jin2021graph} were the first to adapt dataset condensation to the graph domain by proposing a bi-level graph condensation method. Recent advancements in graph condensation have introduced diverse methodologies, targeting both node-level and graph-level tasks. Zheng et al.\cite{zheng2023structure} proposed a structure-free approach that condenses a graph into node embeddings by employing training trajectory matching. Similarly, Liu et al.\cite{liu2022graph} advanced node-level graph condensation by utilizing receptive field distribution matching. \wl{To enhance the efficiency of graph condensation, Xiao et  al.\cite{xiao2024simple} propose the use of a pre-trained Simplified Graph Convolution model. This approach aligns condensed and original graphs, achieving up to a tenfold improvement in computational efficiency while preserving performance, and avoiding the introduction of additional parameters. In a related development, DisCo~\cite{xiao2025disentangled}, a GNN-free graph condensation framework, decouples node and edge condensation into two independent stages, facilitating computational acceleration while achieving accuracy comparable to GNN-based methods.} Neural tangent kernels have also been explored in this domain; for instance, GC-SNTK~\cite{wang2024fast} leverages the Kernel Ridge Regression framework with a structure-based neural tangent kernel to enhance condensation efficiency. \wl{Gao et al.\cite{gao2025rethinking} propose a training-free framework that simplifies graph condensation into a clustering-based class-to-node partition problem.} Furthermore, Zhang et al.\cite{zhang2024navigating} introduce a lossless node-level graph condensation method grounded in trajectory matching. For graph-level condensation, probabilistic modeling of synthetic graph structures is investigated. DosCond~\cite{jin2022condensing} employs a probabilistic approach to model the discrete structure of graphs. In addition, Kernel Ridge Regression-based techniques have been applied to this task. The KiDD method~\cite{xu2023kernel} utilizes the KRR framework to condense entire graph-level datasets effectively.


\section{Preliminary}

Generally, the condensed synthetic graph data $G^{\mathcal{S}}$ is expected to perform as well as the target graph $G$ when applied to the downstream tasks (e.g., training a GNN model $f_{\theta}$ with parameter $\theta$). The current approach to the graph condensation problem involves a bi-level optimization process, where two nested training loops are utilized to optimize the model parameter $\theta$ and the condensed graph $G^{\mathcal{S}}$, respectively. The inner loop handles the training of GNN $f_{\theta_t}$ on the condensed data $G^{\mathcal{S}}$ by the training loss $\ell(f_{\theta}, G^{\mathcal{S}})$, while the outer loop optimizes the condensed graph $G^{\mathcal{S}}$ by minimizing the matching loss $\mathcal{L}(f_{\theta_t}, G)$. Therefore, the bi-level graph condensation problem could be modeled as
\begin{align}\label{DD}
\begin{gathered}
         \mathop{ \min}_{ G^{\mathcal{S}}} \mathcal{L}(f_{\theta}, G ) \\
        s.t. \quad \theta = \mathop{\arg \min}_{\theta} \ell(f_{\theta}, G^{\mathcal{S}}).
\end{gathered}
\end{align}

In general, the convergence of the parameter $\theta$ is influenced by its initial values $\theta_{0}$. This implies that the condensed data will only yield favorable results when the neural network model is initialized with $\theta_{0}$. However, the objective of condensation is to produce data that can perform well under a random initialization distribution $P_{\theta_{0}}$. To address this constraint, multiple random initializations are required, and problem ~\eqref{DD} is accordingly modified as follows:
\begin{align}\label{DD-Exp}
\begin{gathered}
      \mathop{ \min}_{ G^{\mathcal{S}}} \mathbb{E}_{\theta_{0} \sim P_{\theta_{0}}} \left [  \mathcal{L}(f_{\theta}, G )  \right ]  \\
    s.t. \quad \theta = \mathop{\arg \min}_{\theta} \ell(f_{\theta},G^{\mathcal{S}}),
\end{gathered}
\end{align}
where the loss function of the outer loop $\mathcal{L}$ is designed as matching the gradients~\cite{zhao2020dataset} that are calculated by $G^{\mathcal{S}}$ and $G$ on model $f_{\theta_t}$, respectively. This can be expressed as follows:
\begin{align}\label{DD-Loss}
    \mathcal{L} = D(\nabla_{\theta_{t}} \ell(f_{\theta_t}, G^{\mathcal{S}}), \nabla_{\theta_{t}} \ell(f_{\theta_t},G) ),
\end{align}
where $D(\cdot,\cdot)$ is the function measuring the distance between the gradients $\nabla_{\theta_{t}} \ell(f_{\theta_t},G^{\mathcal{S}})$ and $\nabla_{\theta_{t}} \ell(f_{\theta_t},G)$. $\theta_{t}$ is the parameter of the inner loop model updated $t$ times by a specific optimization algorithm $\texttt{opt-alg}_{\theta}$ (e.g., Stochastic Gradient Descent, SGD).
Denote $T$ as the total number of iterations of the outer loop. Finally, the bi-level graph condensation problem becomes:
\begin{align}\label{eq:GCond}
\begin{gathered}
        \mathop{ \min}_{ G^{\mathcal{S}}} \mathbb{E}_{\theta_{0} \sim P_{\theta_{0}}} \left [  \sum_{t=0}^{T} D(\nabla_{\theta_{t}} \ell(f_{\theta_t},G^{\mathcal{S}}), \nabla_{\theta_{t}} \ell(f_{\theta_t},G) )  \right ] \\
        s.t. \quad \theta_{t} = \texttt{opt-alg} \left [\ell(f_{\theta_{t-1}},G^{\mathcal{S}}) \right ].
\end{gathered}
\end{align}

Problem~\eqref{eq:GCond} offers a potential solution for graph data condensation. However, it presents a highly complex optimization challenge. Since a GNN is a parameterized model, the inner loop of this condensation framework requires iterative optimization of the parameters $f_{\theta}$. 
Simultaneously, the outer loop optimizes the condensed data $G^{\mathcal{S}}$. To ensure the generalization capability of the condensed data under initialization distribution $P_{\theta_0}$ of the GNN parameters, an additional outermost loop repeatedly initializes the GNN parameters $\theta_0$, typically exceeding 1,000 iterations. Consequently, this three-level nested optimization algorithm is computationally expensive and time-intensive.

\section{Methodology}

\wl{This section introduces the proposed GCGP framework, which optimizes a synthesized condensed graph specifically designed to enhance the efficiency of downstream tasks. A novel covariance function tailored for graph-structured data serves as the foundation of the framework. To further refine the structural information of the condensed graph, we adopt concrete random variables, which replace the binary adjacency matrix with a differentiable counterpart. Finally, we outline the optimization process and provide a complexity analysis, highlighting the condensation efficiency advantages of the proposed GCGP.}

A target graph dataset $G = {X, A}$ consists of $n$ nodes, where $X \in \mathbb{R}^{n \times d}$ represents the node features of dimensionality $d$, and $A \in {0,1}^{n \times n}$ is the adjacency matrix. Here, $A_{ij} = 1$ indicates a connection between nodes $i$ and $j$, while $A_{ij} = 0$ indicates no connection. Additionally, let $Y$ represent the associated node labels.
Graph condensation is a technique designed to reduce the size of a graph dataset while retaining its critical properties for downstream tasks. Specifically, this method condenses $G$ into a smaller synthetic graph $G^{\mathcal{S}} = { X^{\mathcal{S}}, A^{\mathcal{S}} }$, accompanied by corresponding labels $Y^{\mathcal{S}}$. In the condensed graph, $X^{\mathcal{S}} \in \mathbb{R}^{m \times d}$, where the number of nodes $m$ satisfies $m \ll n$.

\begin{figure*}
    \centering
    \includegraphics[width=0.7\linewidth]{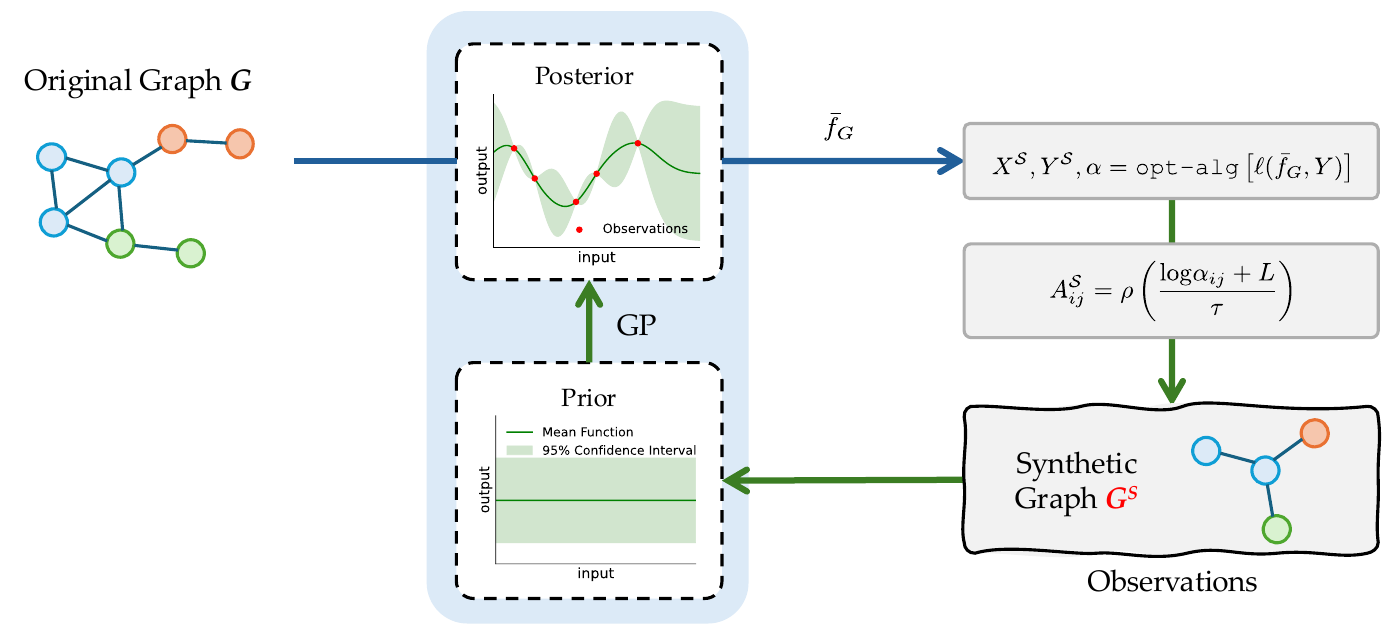}
    \caption{The workflow of the proposed GCGP framework involves three key steps. First, the condensed synthetic graph $G^{\mathcal{S}}$ is utilized as the observations for the GP. Next, predictions are generated for the test locations, corresponding to the original graph $G$. Finally, the condensed graph is iteratively optimized by minimizing the discrepancy between the GP's predictions and the ground-truth labels.}
    \label{fig:gcgp}
\end{figure*}

\subsection{Efficiency Graph Condensation via Gaussian Process}
\wl{Gaussian processes (GPs) provide a probabilistic framework for modeling distributions over functions. Formally, a GP is defined as a collection of random variables, any finite subset of which follows a joint Gaussian distribution~\cite{williams2006gaussian}, which can be defined by a mean function and a covariance function. Usually, for notational simplicity, we will take the mean function to be zero ${\mu}({x}) = 0 $. Given the input $\mathcal{X}$ and the observed target $\mathcal{Y}$, let the test data be $x_t$, the function value be $f_t$, $\mathcal{Y}$ and $f_t$ follow a joint Gaussian process,
\begin{align}\label{eq:prior}
\begin{bmatrix}
{\mathcal{Y}}\\
f_{t}
\end{bmatrix} \sim \mathcal{N} \left ({0},\begin{bmatrix}
\mathcal{K}({\mathcal{X}}, {\mathcal{X}}) + \sigma^2_{\varepsilon}{I} &\mathcal{K}({\mathcal{X}}, x_t)  \\
\mathcal{K}(x_t, {\mathcal{X}}) &\mathcal{K}(x_t, x_t) \end{bmatrix}\right ).
\end{align}

In the context of graph condensation tasks, the condensed data $G^{\mathcal{S}} = {X^{\mathcal{S}}, A^{\mathcal{S}}}$ is treated as a potential input, while the corresponding labels $Y^{\mathcal{S}}$ are considered the observed targets.
We define a prior joint distribution over the observed targets $Y^{\mathcal{S}}$ and the function values $f_G$ at the test locations of the original graph data $G$.
\begin{align}\label{eq:prior}
\begin{bmatrix}
Y^{\mathcal{S}}\\
f_{G}
\end{bmatrix} \sim \mathcal{N} \left ({0},\begin{bmatrix}
\mathcal{K}(G^{\mathcal{S}}, G^{\mathcal{S}}) + \sigma^2_{\varepsilon}{I} &\mathcal{K}(G^{\mathcal{S}}, G)  \\
\mathcal{K}(G, G^{\mathcal{S}}) &\mathcal{K}(G, G) \end{bmatrix}\right ),
\end{align}
where $\mathcal{K}$ is the covariance function, modeling the dependencies between samples in a probabilistic way. In Gaussian processes, the choice of the covariance function is crucial, as it encodes prior assumptions about the function’s smoothness, variability, and how closely the function values at different input points are correlated.

To better understand the origin of the covariance function, we consider a Bayesian linear regression model defined in a high-dimensional feature space. Specifically, denoting a mapping $\phi: \mathbb{R}^d \to \mathbb{R}^{d'}$, which transforms an input ${x}$ into a $d'$-dimensional feature space. Based on this mapping model, a Bayesian linear regression model can be defined as $f({x}) = \phi({x}){W}$, where the weight ${W}$ is assigned a Gaussian prior ${W} \sim \mathcal{N}({0}, {\Sigma}_p)$. In practical scenarios, the true function values are often not directly observable. Instead, we observe noisy outputs $y = f(x) + \varepsilon$, where the noise $\varepsilon$ is assumed to follow an independent and identically distributed Gaussian distribution, $\varepsilon \sim \mathcal{N}(0, \sigma^2_{\varepsilon})$.
This leads to the definition of a covariance function, which encodes input similarity and models dependencies in Gaussian processes,

\begin{align}\label{eq:covy}
    Cov(Y^{\mathcal{S}})    & =    \mathbb{E}[(f(G^{\mathcal{S}})+{\varepsilon})(f(G^{\mathcal{S}})+{\varepsilon})] \notag \\
                    & =    \phi(G^{\mathcal{S}})^{\top} {\Sigma}_p \phi(G^{\mathcal{S}}) + \sigma_{\varepsilon}^2{I} \\
                    & =     \mathcal{K}(G^{\mathcal{S}},G^{\mathcal{S}}) + \sigma^2_{\varepsilon}{I}, \notag
\end{align}
where $I$ is the identity matrix.

The joint prior in Equation~\eqref{eq:prior} defines the distribution over both observed and unobserved function values in Gaussian process regression. By conditioning this prior on the observed data using the standard Gaussian conditioning identity~\cite{williams2006gaussian}, we obtain the posterior distribution over the unobserved points,
\begin{align}
f_{G}|G, G^{\mathcal{S}}, Y^{\mathcal{S}} \sim \mathcal{N}( \bar{f}_G, Cov(f_{G})),
\end{align}
where,
\begin{align}
\bar{f}_G = &  \mathcal{K}(G,G^{\mathcal{S}}) [\mathcal{K}(G^{\mathcal{S}},G^{\mathcal{S}})+\sigma^2_{\varepsilon}{I}]^{-1}Y^{\mathcal{S}},  \label{eq:f} \\
Cov(f_G) = &\mathcal{K}(G,G) -   \notag \\ 
&  \mathcal{K}(G, G^{\mathcal{S}})[\mathcal{K}(G^{\mathcal{S}},G^{\mathcal{S}})+\sigma^2_{\varepsilon}{I}]^{-1}\mathcal{K}(G^{\mathcal{S}},G).  
\end{align}

The predictive mean $\bar{f}_G$ represents the best estimate (in the mean-square sense) of the function at the test inputs, and the predictive variance $\mathrm{Cov}(f_G)$ quantifies the uncertainty associated with these predictions.

}
Traditional bi-level graph condensation methods typically utilize parametric predictive models, such as GNNs, to evaluate the performance of condensed data on tasks like node classification. These methods require iterative training to optimize model parameters, which constitutes the inner-loop optimization in bi-level graph condensation. In contrast, the proposed framework leverages GP, a non-parametric probabilistic model, to make predictions without the need for iterative training. By specifying a joint Gaussian distribution, GP enables conditional inference to estimate function values at unseen data points $G$ while simultaneously quantifying predictive uncertainty~\cite{mackay1998introduction, damianou2013deep}. This design avoids the computational overhead of iterative optimization and parameter initialization inherent in GNN-based methods. As a result, the computational efficiency of GP makes it a strong candidate for graph condensation, enabling the bi-level condensation process to be reformulated as a simpler optimization problem:
\begin{align}\label{eq:gploss}
    \mathop{ \min}_{ G^{\mathcal{S}}} \ell(\bar{f}_G, Y).
\end{align}

Compared to the bi-level condensation model in Equation \eqref{eq:GCond}, this approach simplifies the optimization problem by reformulating it into a single-level optimization for the graph condensation objective. Replacing the GNN with a GP eliminates the iterative training step in the inner loop of traditional bi-level formulations, significantly reducing time complexity and computational costs. This reduction is particularly beneficial for large-scale graph datasets, where the cost of iterative training can become prohibitive. Consequently, the reformulated approach improves scalability and computational efficiency, making it more suitable for real-world applications involving large graphs.


\subsection{Covariance Function for Graph Structural Data}
The covariance function is a fundamental component of GPs because it defines the correlation structure that determines key properties of the posterior distribution, such as smoothness and uncertainty~\cite{neal2012bayesian}. Its design directly influences the predictive model's fitting capability, generalization performance, and computational efficiency, as these aspects are encapsulated in the properties of the resulting covariance matrix~\cite{bishop2006pattern}.

In the context of the graph condensation task, the design of the covariance function is crucial, as it must meet task-specific requirements to effectively extract relevant information while ensuring computational efficiency. As shown in Equation~\eqref{eq:covy}, the covariance function $\mathcal{K}(G^{\mathcal{S}}, G^{\mathcal{S}}) = \phi(G^{\mathcal{S}})^{\top} {\Sigma}_p \phi(G^{\mathcal{S}})$ depends on the mapping model $\phi(\cdot)$ and the matrix $\Sigma_p$. Since the initialization of the weights is sampled from the same distribution, $\Sigma_p$ remains fixed. Therefore, the selection and design of the mapping model $\phi(\cdot)$ play pivotal roles in determining the behavior of the covariance function.

To address the challenges of designing effective mappings for graph data, we propose a computationally efficient mapping model that captures structural information through message passing. The structural information of a graph, which encodes relationships among nodes, is essential for generating meaningful node representations. Our model aggregates $k$-hop neighborhood information in a single step, allowing node features to encompass the $k$-hop receptive field~\cite{wu2019simplifying}. This approach enhances the representation of local graph structures and mitigates the computational overhead associated with multi-step message passing.
The message-passing mechanism is detailed below:
\begin{align}\label{eq:gc}
\hat{X} = \hat{A}^{k}{X}, \notag
\end{align}
where $\hat{A} = \tilde{D}^{-\frac{1}{2}}\tilde{A} \tilde{D}^{-\frac{1}{2}}$. $\tilde{A} = A + I$ is the self-looped adjacency matrix, and $\tilde{D} = \mathrm{diag}(\sum_{j}{\tilde{A}_{1j}}, \sum_{j}{\tilde{A}_{2j}}, \ldots, \sum_{j}{\tilde{A}_{nj}})$.
In the development of models prioritizing simplicity and low computational complexity, node features are updated through a $k$-hop message passing mechanism. This process is followed by a single non-linear operation to improve the expressive capacity of the mapping while minimizing additional computational overhead. Formally, this mapping procedure in Equation~\eqref{eq:covy} can be represented as:
\begin{align}
\phi(X) = \psi(\hat{A}^{k}{X}W) + \varepsilon, 
\end{align}
where $\psi(x) = \frac{2}{\pi}\int_{0}^{x} e^{-t^2} dt$ represents the non-linear activation function. $W$ is the weight matrix.
Building on the mapping, the prediction could be written as:
\wl{\begin{align}\label{eq:y}
Y = & \phi(X)W + \varepsilon, \notag \\
 = & \psi(\hat{A}^{k}{X}W^{(0)} + {\varepsilon}^{(0)})W^{(1)} + {\varepsilon}^{(1)},
\end{align}}
where weight matrices $W^{(0)} \in \mathbb{R}^{d \times d_1}$ and $W^{(1)} \in \mathbb{R}^{d_1 \times C}$ are initialized such that $W^{(0)}, W^{(1)} \sim \mathcal{N}(0, \sigma_w^2)$, $C$ represents the number of possible categories of the input data. The bias terms follow $\varepsilon^{(0)}, \varepsilon^{(1)} \sim \mathcal{N}(0, \sigma_{\varepsilon}^2)$.

\wl{As the width of the hidden layer approaches infinity ($d_1 \to \infty$), the neural network converges to a Gaussian process, enabling an analytical formulation of the covariance function. Assuming that the elements $\hat{x}_i$ in $\hat{X}$ are independent and identically distributed (i.i.d.), the covariance function can be expressed  as the expectations of the post-activation outputs~\cite{neal2012bayesian,lee2017deep}. Let $\beta = \sigma^2_{\varepsilon}$. For two graphs $G = \{X, A\}$ and $G' = \{X', A'\}$, the covariance function $\mathcal{K}$ is defined as:
\begin{align}
    {\mathcal{K}}(G,G') &=  \mathbb{E}_{f \sim \mathcal{GP}(0,\Lambda^{(0)})}   [\psi(f({G}))\psi(f(G'))] + \beta,
\end{align}
where the covariance matrix $\Lambda(G,G')$ is constructed from a base covariance function $\Sigma(G,G')$,  given by,
\begin{align}
    \Lambda(G,G') & = 
        \begin{bmatrix}
            \Sigma(G,G) &\Sigma(G,G') \\
            \Sigma(G',G)&\Sigma(G',G')
        \end{bmatrix}, \\
    {\Sigma}(G,G') &= \frac{1}{d}\hat{A}^{k}{X} ({\hat{A'}^{k}{X'}})^{\top} + \beta,
\end{align}
where $\hat{A}^{k}$ and $\hat{A'}^{k}$ denote the normalized adjacency matrices raised to the $k$-th power, capturing the graph structural information up to  $k$-hop neighborhoods. The term $\beta$ represents the observation noise variance and is incorporated into the predictive covariance to model measurement uncertainty.

The elements of the covariance function $\mathcal{K}(G,G')$ can be calculated analytically as~\cite{williams1996computing},
\begin{align}
{\mathcal{K}}_{ij} = \frac{2}{\pi} \sin^{-1}\frac{2\tilde{x}^{\top}i\Sigma{w} \tilde{x}_j }{\sqrt{(1+2\tilde{x}^{\top}_i\Sigma_w \tilde{x}_i)(1+2\tilde{x}^{\top}_j\Sigma_w \tilde{x}_j)} }, \label{eq:cov}
\end{align}
where $\tilde{x}_i = (\hat{x}_i, 1)^{\top}$ denotes the augmented vector, with $\hat{x}_i$ representing the $i$-th element of $\hat{X}$. The covariance matrix $\Sigma_w = \text{diag}(\sigma_w^2, \sigma_w^2, \cdots, \beta)$ characterizes the noise structure in the model, where $\sigma_w^2$ is typically set to 1.
Equation~\eqref{eq:cov} facilitates computational efficiency by enabling element-wise calculations, which are well-suited for acceleration using GPUs. Thus, this approach provides a computationally efficient and scalable solution for reducing the time cost of GP.
}

\subsection{Learning the Discrete Synthetic Graph Structure}
In the synthetic condensed graph $G^{\mathcal{S}}$, the structural information is encoded in the binary adjacency matrix $A^{\mathcal{S}} \in \{0,1\}^{m \times m}$. 
However, optimizing the adjacency matrix is challenging because it is discrete and typically cannot be directly optimized using gradient-based methods.~\cite{avin2008distance, jin2022condensing}. To address this issue, this work adopts concrete random variables to relax the adjacency matrix, enabling smoother optimization processes and better computational efficiency.


Particularly, we propose applying the Concrete relaxation~\cite{maddison2016concrete, jang2016categorical} technique to approximate the discrete adjacency matrix with a continuous counterpart. For the synthetic adjacency matrix, the Concrete relaxation reformulates the binary variable $A^{\mathcal{S}}{ij}$, which originally follows a Bernoulli distribution, into a differentiable function. This reformulation introduces concrete random variables $\alpha{ij}$ and $\tau$, such that $A^{\mathcal{S}}{ij} \sim \mathrm{BinConcrete}(\alpha{ij}, \tau)$.
Let $L \sim \mathrm{Logistic}$,
$A^{\mathcal{S}}{ij}$ is reparameterized as,
\wl{
\begin{align}\label{eq:adj}
\begin{gathered}
A^{\mathcal{S}}{ij} = \rho \left (\frac{\mathrm{log} \alpha_{ij} + L }{\tau} \right ),
\end{gathered}
\end{align}
where $\rho(x) = \frac{1}{1 + \mathrm{exp}(-x)}$, and $L$ is defined as $L \overset{d}{=} \mathrm{log}(U) - \mathrm{log}(1-U)$, with $U \sim \mathcal{U}(0,1)$.} The parameters $\alpha_{ij}$ and $\tau$ are constrained to the interval $(0, \infty)$, where $\tau$ denotes the temperature parameter.

In Equation~\eqref{eq:adj}, the binary adjacency matrix is reformulated into a differentiable representation parameterized by $\alpha$, facilitating gradient-based optimization. During this optimization process, the temperature parameter $\tau$ is gradually reduced toward zero. This systematic reduction drives the adjacency matrix toward a binary state, effectively approximating its discrete structure. The probability that a specific element of the condensed adjacency matrix, $A^{\mathcal{S}}_{ij}$, equals 1 is:
\begin{align}
\mathbb{P}\left(\lim_{\tau \to 0} A^{\mathcal{S}}_{ij} = 1\right) = \frac{\alpha_{ij}}{1 + \alpha_{ij}}.
\end{align}

To ensure that $A^{\mathcal{S}}_{ij} = 1$, we set the threshold $\mathbb{P}\left(\lim_{\tau \to 0} A^{\mathcal{S}}_{ij} = 1\right) > 0.5$.


\subsection{Objective and Optimization}
This paper introduces GCGP, a novel method that employs GP to integrate prior information with the condensed graph $G^{\mathcal{S}}$ for predictions. The approach is computationally efficient while avoiding additional training loops. The loss function is:
\begin{align}
\begin{gathered}
\ell (\bar{f}{G}, Y) = ||\bar{f}{G} - Y||^{2}_{F}.
\end{gathered}
\end{align}

The optimization process in GCGP involves three key parameters: $X^{\mathcal{S}}$, $Y^{\mathcal{S}}$, and $A^{\mathcal{S}}$. Since $A^{\mathcal{S}}$ is a differentiable function of $\alpha$, the optimization primarily focuses on $X^{\mathcal{S}}$, $Y^{\mathcal{S}}$, and $\alpha$, utilizing algorithms such as stochastic gradient descent. The overall procedure is outlined as follows.

First, the initialization of $X^{\mathcal{S}}$, $Y^{\mathcal{S}}$, $\alpha$, and the adjacency matrix $A^{\mathcal{S}}$ is performed, where $A^{\mathcal{S}}$ is sampled using Equation~\eqref{eq:adj}. Subsequently, GP predictions $\bar{f}_{G}$ for the graph $G$ are computed as described in Equation~\eqref{eq:f}. These predictions are then compared with the original labels $Y$ to calculate the condensation loss. This loss function guides the iterative optimization process, during which $X^{\mathcal{S}}$, $Y^{\mathcal{S}}$, and $\alpha$ are updated.

\wl{To enable a smooth transition from soft to discrete adjacency representations, the temperature parameter \( \tau \) is annealed during training. As \( \tau \to 0 \), the sampled adjacency matrix \( A^{\mathcal{S}} \) becomes increasingly binary. The annealing schedule is defined as:
\begin{align}
    \tau = \max\left( \text{temp}_{\text{start}} \cdot \left( \frac{\text{temp}_{\text{end}}}{\text{temp}_{\text{start}}} \right)^{\frac{\text{epoch}}{100}},\ \text{temp}_{\text{end}} \right),
\end{align}
where \( \text{temp}_{\text{start}} = 0.3 \) and \( \text{temp}_{\text{end}} = 0.01 \). The full training procedure is shown in Algorithm~\ref{alg}.}


\begin{algorithm}[t]
	\caption{{GPGC}}
    \label{alg}
	\begin{algorithmic}[1]
        \STATE \textbf{Input:} Target graph data $G = \{ X,A \}$ with $n$ nodes and labels $Y$.\\
        \STATE \textbf{Output:} Condensed graph data $G^{\mathcal{S}} = \{ X^{\mathcal{S}},A^{\mathcal{S}} \}$ with $m (m \ll n)$ nodes and labels $Y^{\mathcal{S}}$. \\
        \STATE \wl{\textbf{Initialize:} $ X^{\mathcal{S}}_{ij}, Y^{\mathcal{S}}_{ij}\sim \mathcal{U}(0,1), \quad \forall i,j$, and $ \alpha_{ij} \sim \mathcal{U}(0,2), \quad \forall i,j $.}\\

        \WHILE{not converge}
        \STATE Update $A^{\mathcal{S}}$ by:
        \wl{$$ A^{\mathcal{S}}_{ij} = \rho \left (\frac{\mathrm{log} \alpha_{ij} + L }{\tau} \right ) $$}
        \STATE Calculate $\mathcal{K}(G,G^{\mathcal{S}})$ and $\mathcal{K}(G^{\mathcal{S}},G^{\mathcal{S}})$ by Equation~\eqref{eq:cov}. \\
        
        \STATE Calculate loss function:
        $$ \ell (\bar{f}_{G}, Y) = ||\mathcal{K}(G,G^{\mathcal{S}}) [\mathcal{K}(G^{\mathcal{S}},G^{\mathcal{S}})+\beta{I}]^{-1}Y^{\mathcal{S}} - Y||^{2}_{F} $$ \\
        \STATE Update $X^{\mathcal{S}}, Y^{\mathcal{S}}, \alpha$:
        \[ X^{\mathcal{S}}, Y^{\mathcal{S}}, \alpha = \texttt{opt-alg} \left [\ell (\bar{f}_{G},Y)\right ]\] \\
        \STATE Update $ \tau \to 0 $ \\
        \ENDWHILE
        \IF{ $\frac{\alpha_{ij}}{1 + \alpha_{ij}} > 0.5$ }
        \STATE $A^{\mathcal{S}}_{ij} = 1$
        \ELSE 
        \STATE $A^{\mathcal{S}}_{ij} = 0$
        \ENDIF
	\end{algorithmic}
\end{algorithm}

\subsection{Complexity Analysis}

To demonstrate that the proposed GCGP method has lower computational complexity than the bi-level method GCond, we perform a detailed complexity analysis. The notations are defined as follows: $n$ and $m$ denote the numbers of nodes in the original and condensed datasets, respectively, $d$ is the dimensionality of node features, and $|E|$ represents the number of edges in the original graph. Additionally, $t_{in}$ and $t_{out}$ are the numbers of iterations in the inner and outer loops, respectively, while $t_{init}$ denotes the number of times the GNN model is re-initialized in GCond.

The computational complexity of GCond is derived by analyzing its inner and outer loops. The inner loop, involving GNN training, has a complexity of $O(t_{in}(md^2))$. For the outer loop, the primary cost is forward propagation on the original dataset, with a complexity of $O(nd^2 + |E|d)$. Considering $t_{out}$ optimization iterations, the total outer loop complexity becomes $O(t_{out}(nd^2 + |E|d))$. Combining these, the overall complexity of GCond is:$ O(t_{init}(t_{out}(nd^2 + |E|d + t_{in}(md^2)))$.

The proposed GCGP method involves two main steps: covariance function computation and GP prediction. The covariance computation has a complexity of $O(mnd)$, while GP prediction has $O(m^3 + mnd)$. Thus, the total complexity of GCGP is: $O(tm^3 + tmnd)$.
This analysis highlights that the GCGP method achieves significantly lower computational complexity than GCond, particularly for large-scale datasets where $n$ and $|E|$ dominate the cost.
\section{Experiments}

{
\setlength{\tabcolsep}{1pt}
\begin{table}[]
\centering
\caption{Dataset details}
\label{tab:data}
\renewcommand{\arraystretch}{1.2}
\begin{tabular}{crrrrr}
\hline
\textbf{Dataset} & \multicolumn{1}{c}{\textbf{Nodes}} & \multicolumn{1}{c}{\textbf{Edges}} & \multicolumn{1}{c}{\textbf{Classes}} & \multicolumn{1}{c}{\textbf{Features}} & \multicolumn{1}{c}{\textbf{Train/Validation/Test}} \\ \hline
Cora             & 2,708                              & 5,429                              & 7                                    & 1,433                                 & 140/500/1,000                                      \\
Citeseer         & 3,327                              & 9,104                              & 6                                    & 3,703                                 & 120/500/1,000                                      \\
Pubmed           & 19,717                             & 44,338                             & 3                                    & 500                                   & 60/500/1,000                                       \\
Photo            & 7,650                              & {\color[HTML]{404040} 238,162}     & 8                                    & 745                                   & 160/500/800                                        \\
Computers        & 13,752                             & 491,722                            & 10                                   & 767                                   & 200/500/1,000                                      \\
Ogbn-arxiv       & 169,343                            & 1,166,243                          & 40                                   & 128                                   & 90,941/29,799/48,603                               \\
Reddit           & 232,965                            & 114,615,892                        & 41                                   & 602                                   & 153,431/23,831/55,703                              \\ \hline
\end{tabular}
\end{table}
}

In this section, we evaluate the condensation effectiveness of our method across various settings. Extensive comparative experiments demonstrate that our method consistently achieves the fastest performance on both small and large datasets. To assess generalization, we train different neural network models using the condensed data and analyze their performance. A sensitivity analysis of parameters is also conducted. Finally, ablation experiments confirm the method's advantages in efficiency and condensation quality. All experiments are executed on an NVIDIA RTX 3090 GPU with 24GB RAM.

{
\setlength{\tabcolsep}{2pt}
\begin{table}[t]
\centering
\caption{Configuration of the experiments}
\label{tab:config}
\renewcommand{\arraystretch}{1.2}
\begin{tabular}{cccccccccc}
\hline
\textbf{Dataset}          & \textbf{Size} & \textbf{$\beta$} & $k$ & \textbf{Learn A} & \textbf{Dataset}            & \textbf{Size} & \textbf{$\beta$} & ${k}$ & \textbf{Learn A} \\ \hline
\multirow{6}{*}{Cora}     & 35            & 0.01           & 3          & 0                & \multirow{6}{*}{Computers}  & 50            & 0.01           & 1          & 0                \\
                          & 35            & 0.5            & 2          & 1                &                             & 50            & 0.01           & 1          & 1                \\
                          & 70            & 0.5            & 4          & 0                &                             & 100           & 0.01           & 2          & 0                \\
                          & 70            & 1              & 2          & 1                &                             & 100           & 0.01           & 1          & 1                \\
                          & 140           & 0.5            & 4          & 0                &                             & 200           & 0.01           & 2          & 0                \\
                          & 140           & 0.01           & 2          & 1                &                             & 200           & 0.001          & 1          & 1                \\ \hline
\multirow{6}{*}{Citeseer} & 30            & 10             & 5          & 0                & \multirow{6}{*}{Ogbn-arxiv} & 90            & 5              & 2          & 0                \\
                          & 30            & 1              & 2          & 1                &                             & 90            & 0.5            & 2          & 1                \\
                          & 60            & 0.5            & 2          & 0                &                             & 454           & 0.001          & 2          & 0                \\
                          & 60            & 5              & 1          & 1                &                             & 454           & 0.5            & 2          & 1                \\
                          & 120           & 5              & 4          & 0                &                             & 909           & 10             & 2          & 0                \\
                          & 120           & 5              & 1          & 1                &                             & 909           & 0.5            & 2          & 1                \\ \hline
\multirow{6}{*}{Pubmed}   & 15            & 0.001          & 2          & 0                & \multirow{6}{*}{Reddit}     & 77            & 0.1            & 2          & 0                \\
                          & 15            & 0.5            & 2          & 1                &                             & 77            & 0.01           & 2          & 1                \\
                          & 30            & 0.01           & 2          & 0                &                             & 153           & 0.1            & 2          & 0                \\
                          & 30            & 0.5            & 2          & 1                &                             & 153           & 0.1            & 1          & 1                \\
                          & 60            & 0.1            & 5          & 0                &                             & 307           & 1              & 2          & 0                \\
                          & 60            & 0.5            & 2          & 1                &                             & 307           & 0.1            & 2          & 1                \\ \hline
\multirow{3}{*}{Photo}    & 40            & 0.1            & 2          & 0                & \multirow{3}{*}{Photo}      & 80            & 0.01           & 1          & 1                \\
                          & 40            & 0.01           & 1          & 1                &                             & 160           & 0.001          & 2          & 0                \\
                          & 80            & 0.1            & 2          & 0                &                             & 160           & 0.001          & 1          & 1                \\ \hline
\end{tabular}
\end{table}
}

\setlength{\tabcolsep}{5pt}
\begin{table*}[]
\centering
\caption{\wl{The model's average accuracy and standard deviation are assessed based on the condensed graph data. Green shading highlights the top three condensation methods in each row, with darker shades denoting superior performance.}}
\label{tab:acc}
\renewcommand{\arraystretch}{1.3}
\setlength{\tabcolsep}{4pt} 
\begin{tabular}{ccccccccccccc}
\hline
                                                                                  &                                         &                                     & \multicolumn{2}{c}{\textbf{GCond}}                                  & \multicolumn{2}{c}{\textbf{GC-SNTK}}                                &                                  &                                  &                                  & \multicolumn{2}{c}{\textbf{GCGP (Our)}}                             &                                       \\ \cline{4-7} \cline{11-12}
\multirow{-2}{*}{\textbf{Dataset}}                                                & \multirow{-2}{*}{\textbf{Ratio (Size)}} & \multirow{-2}{*}{\textbf{One-step}} & \textbf{X}                       & \textbf{X,A}                     & \textbf{X}                       & \textbf{X,A}                     & \multirow{-2}{*}{\textbf{SFGC}}  & \multirow{-2}{*}{\textbf{SimGC}} & \multirow{-2}{*}{\textbf{DisCo}} & \textbf{X}                       & \textbf{X,A}                     & \multirow{-2}{*}{\textbf{Full (GCN)}} \\ \hline
                                                                                  & 1.30\% (35)                             & 80.2±0.73                           & 75.9±1.2                         & 81.2±0.7                         & \cellcolor[HTML]{D8E4BC}82.2±0.3 & \cellcolor[HTML]{EBF1DE}81.7±0.7 & 80.1±0.4                         & 80.8±2.3                         & 76.9±0.8                         & \cellcolor[HTML]{C4D79B}82.4±0.3 & 78.8±1.0                         &                                       \\
                                                                                  & 2.60\% (70)                             & 80.4±1.77                           & 75.7±0.9                         & 81.0±0.6                         & \cellcolor[HTML]{D8E4BC}82.4±0.5 & 81.5±0.7                         & \cellcolor[HTML]{EBF1DE}81.7±0.5 & 80.9±2.6                         & 78.7±0.3                         & \cellcolor[HTML]{C4D79B}82.5±0.7 & 80.5±0.9                         &                                       \\
\multirow{-3}{*}{Cora}                                                            & 5.20\% (140)                            & 79.8±0.64                           & 76.0±0.9                         & 81.1±0.5                         & \cellcolor[HTML]{D8E4BC}82.1±0.1 & 81.3±0.2                         & 81.6±0.8                         & \cellcolor[HTML]{EBF1DE}82.1±1.3 & 78.8±0.5                         & \cellcolor[HTML]{C4D79B}82.6±0.7 & 80.9±1.0                         & \multirow{-3}{*}{81.1±0.5}            \\ \hline
                                                                                  & 0.90\%(30)                              & 70.8±0.3                            & 71.6±0.8                         & \cellcolor[HTML]{EBF1DE}71.8±1.2 & 65.7±0.3                         & 64.8±0.7                         & 71.4±0.5                         & \cellcolor[HTML]{C4D79B}73.8±2.5 & 67.9±0.7                         & \cellcolor[HTML]{D8E4BC}73.1±1.5 & 70.3±1.5                         &                                       \\
                                                                                  & 1.80\%(60)                              & 71.7±0.6                            & 70.8±0.1                         & \cellcolor[HTML]{D8E4BC}72.6±0.9 & 67.0±0.3                         & 65.9±0.2                         & \cellcolor[HTML]{EBF1DE}72.4±0.4 & 72.2±0.5                         & 67.6±0.5                         & \cellcolor[HTML]{C4D79B}72.8±0.6 & 71.3±1.0                         &                                       \\
\multirow{-3}{*}{CiteSeer}                                                        & 3.61\%(120)                             & 70.1±0.2                            & \cellcolor[HTML]{EBF1DE}71.4±0.6 & \cellcolor[HTML]{C4D79B}72.5±0.4 & 67.3±0.3                         & 66.3±0.5                         & 70.6±0.7                         & 71.1±2.8                         & 67.9±0.9                         & \cellcolor[HTML]{D8E4BC}72.3±0.5 & 71.1±0.5                         & \multirow{-3}{*}{71.7±0.1}            \\ \hline
                                                                                  & 0.08\% (15)                             & 77.7±0.12                           & 59.4±0.7                         & 78.3±0.2                         & \cellcolor[HTML]{D8E4BC}78.9±0.7 & 71.8±6.8                         & \cellcolor[HTML]{EBF1DE}78.7±0.5 & 71.0±1.2                         & 76.1±0.8                         & \cellcolor[HTML]{C4D79B}79.2±0.5 & 74.2±0.7                         &                                       \\
                                                                                  & 0.15\% (30)                             & 77.8±0.17                           & 51.7±0.4                         & \cellcolor[HTML]{FFFFFF}77.1±0.3 & \cellcolor[HTML]{C4D79B}79.3±0.3 & \cellcolor[HTML]{FFFFFF}74.0±4.9 & \cellcolor[HTML]{EBF1DE}78.5±0.7 & \cellcolor[HTML]{FFFFFF}70.8±2.1 & \cellcolor[HTML]{FFFFFF}77.5±0.6 & \cellcolor[HTML]{D8E4BC}79.2±0.7 & \cellcolor[HTML]{FFFFFF}76.8±0.9 &                                       \\
\multirow{-3}{*}{PubMed}                                                          & 0.30\% (60)                             & 77.1±0.44                           & 60.8±1.7                         & \cellcolor[HTML]{FFFFFF}78.4±0.3 & \cellcolor[HTML]{C4D79B}79.4±0.3 & \cellcolor[HTML]{FFFFFF}76.4±2.8 & \cellcolor[HTML]{EBF1DE}78.8±0.8 & \cellcolor[HTML]{FFFFFF}71.0±1.8 & \cellcolor[HTML]{FFFFFF}77.7±0.4 & \cellcolor[HTML]{D8E4BC}79.2±0.8 & \cellcolor[HTML]{FFFFFF}77.5±0.5 & \multirow{-3}{*}{77.1±0.3}            \\ \hline
                                                                                  & 0.5\% (40)                              & 85.7±0.4                            & 86.6±0.6                         & \cellcolor[HTML]{FFFFFF}85.5±0.2 & \cellcolor[HTML]{FFFFFF}88.6±0.1 & \cellcolor[HTML]{FFFFFF}82.0±2.5 & \cellcolor[HTML]{EBF1DE}90.1±0.3 & \cellcolor[HTML]{FFFFFF}89.5±1.3 & \cellcolor[HTML]{FFFFFF}88.1±0.8 & \cellcolor[HTML]{C4D79B}92.1±0.4 & \cellcolor[HTML]{D8E4BC}90.2±0.9 &                                       \\
                                                                                  & 1.0\% (80)                              & 86.2±0.2                            & 87.3±0.4                         & 86.5±0.5                         & 88.5±0.2                         & 83.7±1.0                         & \cellcolor[HTML]{D8E4BC}90.7±0.8 & 89.4±0.9                         & 89.5±0.9                         & \cellcolor[HTML]{C4D79B}92.0±0.1 & \cellcolor[HTML]{EBF1DE}90.6±0.5 &                                       \\
\multirow{-3}{*}{\begin{tabular}[c]{@{}c@{}}Amazon\\      Photo\end{tabular}}     & 2.1\% (160)                             & 86.8±0.6                            & 87.7±0.3                         & 86.6±0.6                         & 88.6±0.6                         & 83.8±3.0                         & \cellcolor[HTML]{EBF1DE}90.6±0.6 & 89.6±1.1                         & 89.3±0.6                         & \cellcolor[HTML]{C4D79B}92.1±0.3 & \cellcolor[HTML]{D8E4BC}91.4±0.6 & \multirow{-3}{*}{91.4±0.8}            \\ \hline
                                                                                  & 0.4\% (50)                              & 82.1±0.3                            & 82.4±0.6                         & 82.7±0.3                         & 83.5±0.3                         & 77.2±5.2                         & \cellcolor[HTML]{EBF1DE}84.7±0.5 & 82.9±1.2                         & 82.4±0.6                         & \cellcolor[HTML]{C4D79B}86.1±0.3 & \cellcolor[HTML]{D8E4BC}85.1±0.5 &                                       \\
                                                                                  & 0.7\% (100)                             & 82.4±0.5                            & 82.2±0.5                         & 82.5±0.2                         & 83.1±0.6                         & 77.3±3.5                         & \cellcolor[HTML]{EBF1DE}85.0±0.2 & 82.6±1.0                         & 82.7±0.4                         & \cellcolor[HTML]{C4D79B}86.4±0.3 & \cellcolor[HTML]{D8E4BC}85.4±0.4 &                                       \\
\multirow{-3}{*}{\begin{tabular}[c]{@{}c@{}}Amazon\\      Computers\end{tabular}} & 1.5\% (200)                             & 82.3±0.6                            & 82.7±0.3                         & 82.8±0.4                         & 83.0±0.7                         & 77.7±5.0                         & \cellcolor[HTML]{EBF1DE}85.5±0.7 & 83.9±0.9                         & 84.0±0.7                         & \cellcolor[HTML]{C4D79B}86.5±0.4 & \cellcolor[HTML]{D8E4BC}86.0±0.5 & \multirow{-3}{*}{84.2±0.3}            \\ \hline
                                                                                  & 0.05\% (90)                             & 59.2±0.1                            & 61.3±0.5                         & 59.2±1.1                         & 63.5±0.3                         & 64.2±0.2                         & \cellcolor[HTML]{D8E4BC}65.5±0.7 & 63.6±0.8                         & 64.0±0.7                         & \cellcolor[HTML]{C4D79B}65.7±0.3 & \cellcolor[HTML]{EBF1DE}65.0±0.4 &                                       \\
                                                                                  & 0.25\% (454)                            & 60.1±0.8                            & 64.2±0.4                         & 63.2±0.3                         & 64.5±0.1                         & 65.1±0.8                         & \cellcolor[HTML]{D8E4BC}66.1±0.4 & \cellcolor[HTML]{C4D79B}66.4±0.3 & \cellcolor[HTML]{EBF1DE}65.9±0.5 & 65.3±0.1                         & 65.4±0.1                         &                                       \\
\multirow{-3}{*}{Ogbn-arxiv}                                                      & 0.5\% (909)                             & 60.0±0.1                            & 63.1±0.5                         & 64.0±0.4                         & 65.7±0.4                         & 65.4±0.5                         & \cellcolor[HTML]{D8E4BC}66.8±0.4 & \cellcolor[HTML]{C4D79B}66.8±0.4 & \cellcolor[HTML]{EBF1DE}66.2±0.1 & 65.7±0.1                         & 65.3±0.3                         & \multirow{-3}{*}{71.4±0.1}            \\ \hline
                                                                                  & 0.05\% (77)                             & 80.7±0.2                            & 88.4±0.4                         & 88.0±1.8                         & 77.9±0.9                         & 74.3±0.5                         & 89.7±0.2                         & 89.6±0.6                         & \cellcolor[HTML]{EBF1DE}91.4±0.2 & \cellcolor[HTML]{C4D79B}93.9±0.1 & \cellcolor[HTML]{D8E4BC}92.7±0.1 &                                       \\
                                                                                  & 0.1\%(153)                              & 81.1±0.4                            & 89.3±0.1                         & 89.6±0.7                         & 78.7±0.6                         & 74.8±0.7                         & 90.0±0.3                         & \cellcolor[HTML]{EBF1DE}92.0±0.3 & 91.8±0.3                         & \cellcolor[HTML]{C4D79B}93.8±0.1 & \cellcolor[HTML]{D8E4BC}92.6±0.1 &                                       \\
\multirow{-3}{*}{Reddit}                                                          & 0.2\% (307)                             & 82.3±0.7                            & 88.8±0.4                         & 90.1±0.5                         & 78.9±0.1                         & 85.2±1.2                         & 89.9±0.4                         & \cellcolor[HTML]{D8E4BC}92.6±0.1 & 91.7±0.3                         & \cellcolor[HTML]{C4D79B}94.0±0.1 & \cellcolor[HTML]{EBF1DE}92.6±0.4 & \multirow{-3}{*}{93.9±0.0}            \\ \hline
\end{tabular}
\end{table*}

\subsection{Experimental Settings}

\subsubsection{Datasets}
\wl{We evaluate on seven benchmark datasets of varying scales: \textbf{Cora}, \textbf{Citeseer}, and \textbf{Pubmed}~\cite{yang2016revisiting} are citation networks with bag-of-words or TF-IDF features~\cite{liu2018research}; \textbf{Photo} and \textbf{Computers}~\cite{shchur2018pitfalls} are Amazon co-purchase graphs with product review features~\cite{qader2019overview}; \textbf{Ogbn-arxiv}~\cite{hu2020ogb} is a citation graph of arXiv CS papers with averaged word embeddings as features; and \textbf{Reddit}~\cite{hamilton2017inductive} is a large-scale social network with post-level community labels. The details of the datasets are shown in Table~\ref{tab:data}.}

\subsubsection{Baseline Methods}
To evaluate the proposed GCGP method comprehensively, we compare it against various graph condensation methods serving as baselines, including,
\begin{itemize}
\item \textbf{One-step}\cite{jin2022condensing} simplifies the GCond method by performing the inner and outer optimization steps only once, thereby accelerating the condensation process.
\item \textbf{GCond}\cite{jin2021graph} is a bi-level graph condensation method that employs a GNN as the prediction model.
\item \textbf{GC-SNTK}\cite{wang2024fast} is a graph condensation method based on kernel ridge regression, utilizing the structure-based neural tangent kernel to quantify similarity among nodes.
\item \textbf{SFGC}\cite{zheng2023structure} is a structure-free graph condensation method that distills large graphs into compact node sets without explicit structural information.
\item \textbf{SimGC}\cite{xiao2024simple} uses a pre-trained SGC model to align condensed and original graphs, achieving up to 10 times speedup with competitive performance.
\item \textbf{DisCo}\cite{xiao2025disentangled} is a GNN-free graph condensation method that decouples node and edge condensation into two stages, achieving speedup and competitive accuracy.
\end{itemize}

We evaluate graph condensation using three methods: GCond, GC-SNTK, and the proposed GCGP, under two distinct scenarios. In the first scenario, the condensed graph contains only node features $X$, while its adjacency matrix $A^{\mathcal{S}}$ is fixed as the identity matrix $I$. In the second scenario, both the node features $X$ and the graph structure $A^{\mathcal{S}}$ are optimized during condensation. These scenarios allow for a comprehensive comparison of the methods' performance under varying levels of structural information.

The GCond framework supports various combinations of GNNs for the condensation and testing stages. By default, the experimental section adopts the best-performing configuration: SGC~\cite{wu2019simplifying} for condensation and GCN~\cite{kipf2016semi} for testing, unless otherwise specified.

\subsubsection{Hyperparameter Settings}
The hyperparameters analyzed in this study include $\beta$ in the GP and the power $k$ of the adjacency matrix $A$ used in the covariance function computation. To ensure consistency across datasets, $\beta$ is assigned the values $\{10^{-3}, 10^{-2}, 10^{-1}, 0.5, 1, 5, 10\}$ for most datasets, except for Reddit, where it is set to $\{10^{-3}, 10^{-2}, 10^{-1}, 1, 10\}$. Similarly, the parameter $k$ is evaluated with $k = \{1, 2, 3, 4, 5\}$ for the Cora, Citeseer, Pubmed, Photo, and Computer datasets, while $k = \{1, 2, 3, 4\}$ is used for the Ogbn-arxiv and Reddit datasets. The detailed hyperparameter configurations and the corresponding optimal values for each dataset are summarized in Table~\ref{tab:config}.

\subsection{Condensation Quality Evaluation}
To assess the quality of synthetic graphs condensed by the proposed GCGP method, we perform node classification experiments across seven datasets, each with varying condensation scales. For the Cora, Citeseer, Pubmed, Photo, and Computers datasets, three condensation scales of 5, 10, and 20 nodes per class are selected. For the Ogbn-arxiv dataset, we use 0.05\% (90 nodes), 0.25\% (454 nodes), and 0.5\% (909 nodes) of the training set as the condensation scales. Similarly, for the Reddit dataset, the condensation scales are set to 0.05\% (77 nodes), 0.1\% (153 nodes), and 0.2\% (307 nodes). \wl{Table~\ref{tab:acc} presents the classification accuracy and standard deviation of various graph condensation methods across multiple datasets and condensation ratios. The top three performing methods for each setting are highlighted in green, with darker shades indicating better performance.}

\wl{Overall, the proposed GCGP method consistently delivers top-tier performance across diverse datasets and condensation ratios. On Cora and Citeseer, it achieves 82.5\% and 72.8\% accuracy at 2.6\% and 1.8\% condensation rates, respectively, outperforming all baselines.}

\wl{On Amazon Photo and Computers, GCGP not only surpasses other condensation methods but also outperforms full-data GCN. For example, with just 0.5\% of nodes on Photo, GCGP achieves 92.1\% accuracy, exceeding the 91.4\% from full GCN. A similar gain is observed on Computers (86.1\% vs. 84.2\%).
On large-scale datasets, GCGP demonstrates scalability and competitive performance. For instance, on Ogbn-arxiv, it achieves 65.7\% accuracy using only 90 condensed nodes. On Reddit, GCGP reaches 93.9\% accuracy with merely 77 synthetic nodes (0.05\% of the original size), closely matching the result obtained by GCN trained on all 153,431 nodes. These results highlight GCGP's effectiveness in extreme compression scenarios and its ability to generalize across diverse graph domains.}

In summary, GCGP outperforms existing baseline methods in graph data condensation, particularly under high condensation ratios and on large-scale datasets. Notably, for six out of the seven datasets (excluding Ogbn-arxiv), the results obtained with condensed data surpass those achieved by training on the full dataset with GCN.

\subsection{Efficient Evaluation}

\begin{figure}[t]
  \centering
  \subfigure[\wl{Cora(70)}]
  {
    \includegraphics[width=0.21\textwidth]{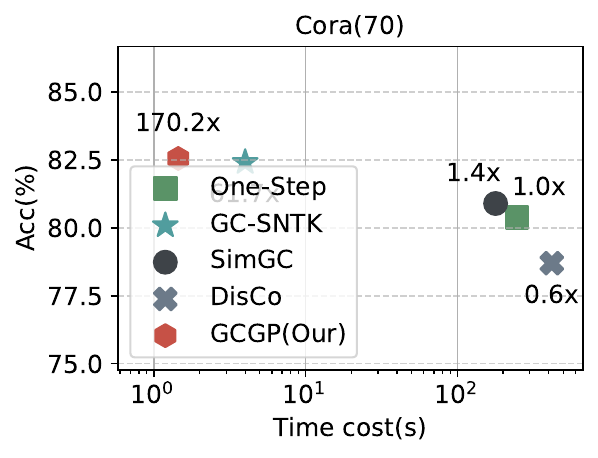}
  }
  \subfigure[\wl{Citeseer(60)}]
  {
    \includegraphics[width=0.21\textwidth]{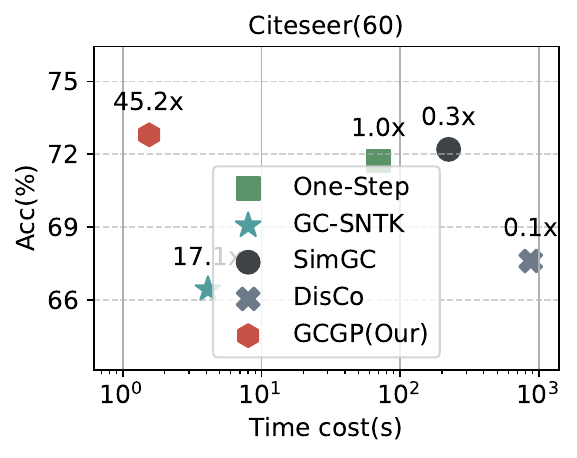}
  }\vspace{-0.2cm}
  
  \subfigure[\wl{Pubmed(30)}]
  {
    \includegraphics[width=0.21\textwidth]{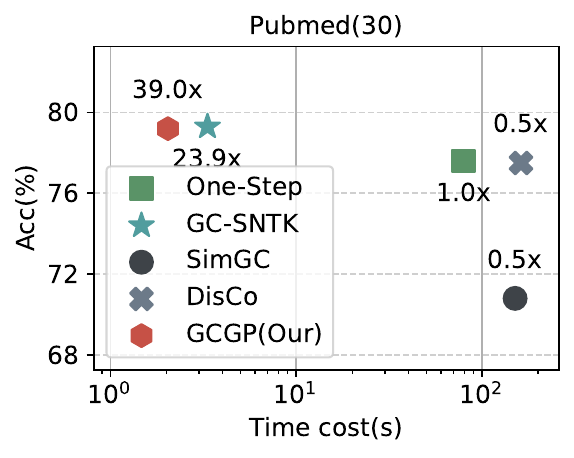}
  }
  \subfigure[\wl{Ogbn-arxiv(90)}]
  {
    \includegraphics[width=0.21\textwidth]{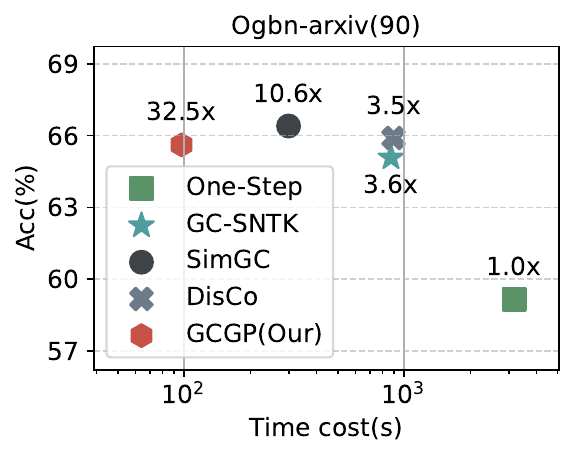}
  }\vspace{-0.2cm}
  \caption{\wl{The runtime efficiency of GC-SNTK, SimGC, DisCo, and GCGP is assessed using the One-step method as the baseline. To quantify relative performance, a speedup factor is calculated by dividing the runtime of the One-step method by the runtime of the other methods and represents the relative acceleration of each method.}}
  \label{fig:log_time}
\end{figure} 

To evaluate the efficiency of the proposed GCGP method, we examine its graph condensation time. The assessment involves a comparison with baseline methods, focusing on computational performance for both small-scale and large-scale graphs. This analysis aims to highlight the scalability and practical applicability of the proposed approach. \wl{Specifically, we compare it with several representative baselines, including One-step, GC-SNTK, SimGC, and DisCo. GCond and SFGC are excluded from this comparison due to their prohibitive computational cost. Experiments are conducted on four benchmark datasets: Cora, Citeseer, Pubmed, and Ogbn-arxiv.}

Figure~\ref{fig:log_time} illustrates the relationship between condensation time (x-axis) and classification accuracy (y-axis), with each marker representing a method’s performance. Using One-step as the baseline, we annotate the relative speedup of each method beside its marker to highlight efficiency gains.
As shown in the figure, GCGP consistently demonstrates superior efficiency while maintaining high accuracy. For instance, on Cora with 70 condensed nodes, GC-SNTK achieves a 61.7$\times$ speedup over One-step, whereas GCGP further improves this to 170.3$\times$. On Citeseer and Ogbn-arxiv, GCGP achieves speedups of 45.2$\times$ and 32.5$\times$, respectively. These results confirm that GCGP accelerates the condensation process without compromising the quality of the condensed graphs.

The results in Figure~\ref{fig:log_time} demonstrate that the proposed GCGP method improves condensation efficiency while maintaining high classification accuracy, outperforming other graph condensation methods. This efficiency is particularly important, as the primary purpose of graph condensation is to accelerate training. If the condensation process itself is excessively time-consuming, it undermines this goal. By substantially reducing condensation time without sacrificing performance, GCGP enhances the practicality and scalability of graph condensation in real-world applications.

\wl{
\subsection{Comparison Between GCGP and Kernel-Based Method}\label{sec:effi}

\begin{figure*}[t]
  \centering
  \subfigure[Cora (35)]
  {
    \includegraphics[width=0.14\textwidth]{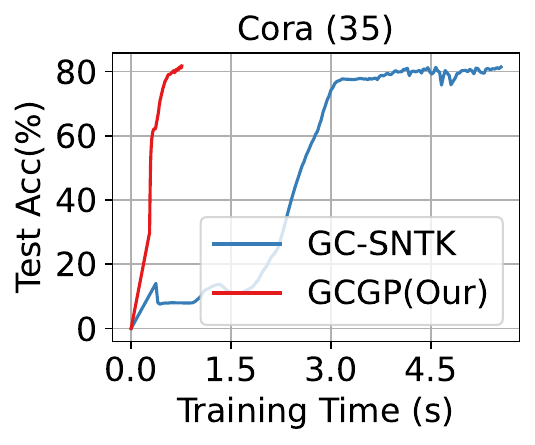}
  }
  \subfigure[Citeseer (30)]
  {
    \includegraphics[width=0.14\textwidth]{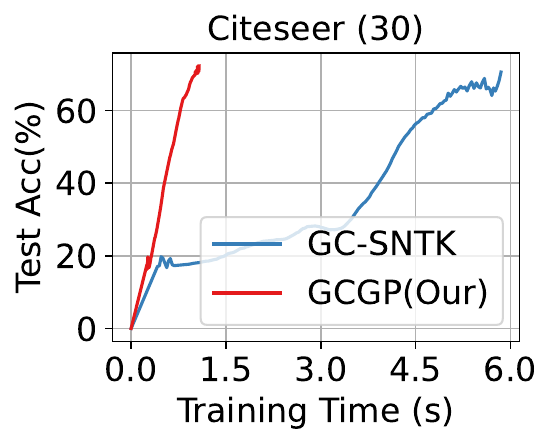}
  }
  \subfigure[Pubmed (15)]
  {
    \includegraphics[width=0.14\textwidth]{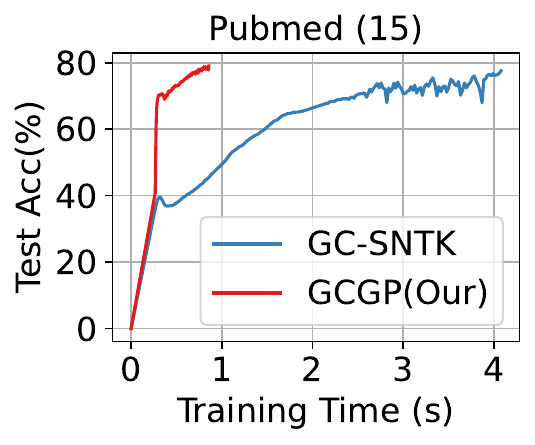}
  }
  \subfigure[Photo (40)]
  {
    \includegraphics[width=0.14\textwidth]{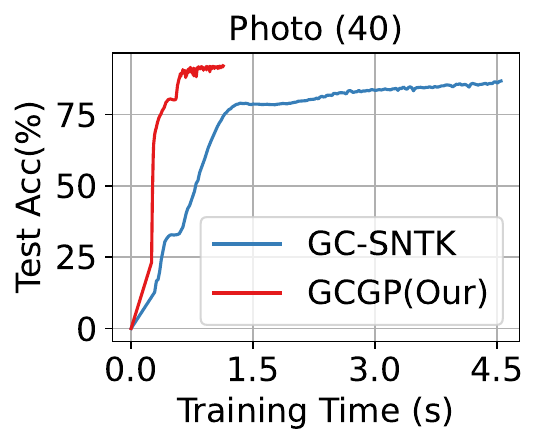}
  }
  \subfigure[Computers (50)]
  {
    \includegraphics[width=0.14\textwidth]{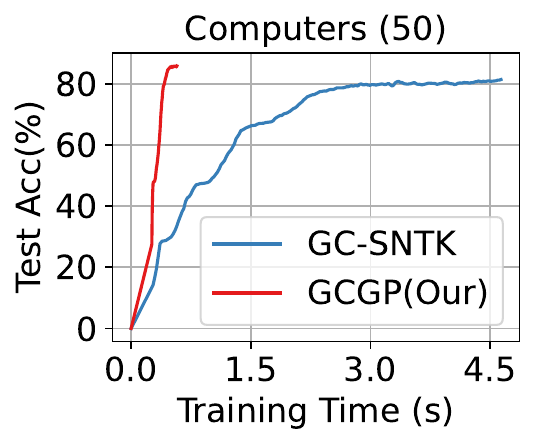}
  }
  \subfigure[\wl{Ogbn-arxiv (90)}]
  {
    \includegraphics[width=0.14\textwidth]{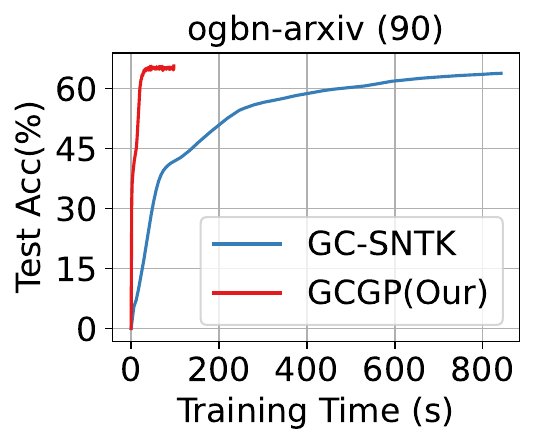}
  }
  
  \subfigure[Cora (70)]
  {
    \includegraphics[width=0.14\textwidth]{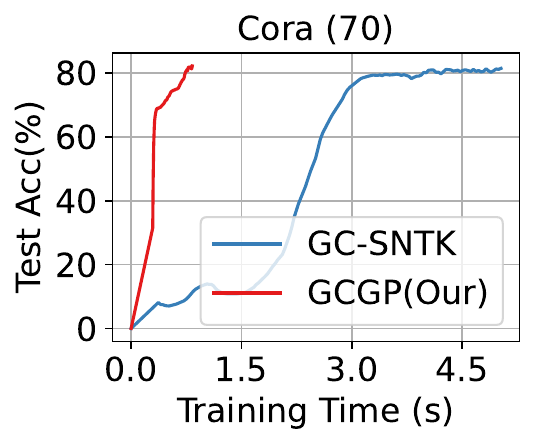}
  }
  \subfigure[Citeseer (60)]
  {
    \includegraphics[width=0.14\textwidth]{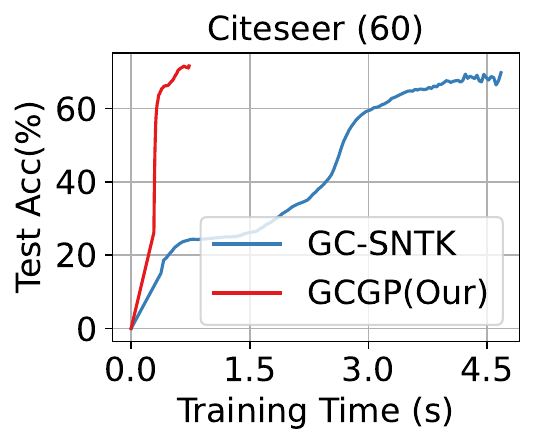}
  }
  \subfigure[Pubmed (30)]
  {
    \includegraphics[width=0.14\textwidth]{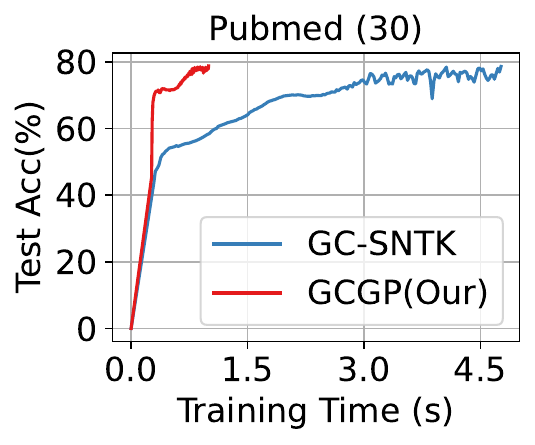}
  }
  \subfigure[Photo (80)]
  {
    \includegraphics[width=0.14\textwidth]{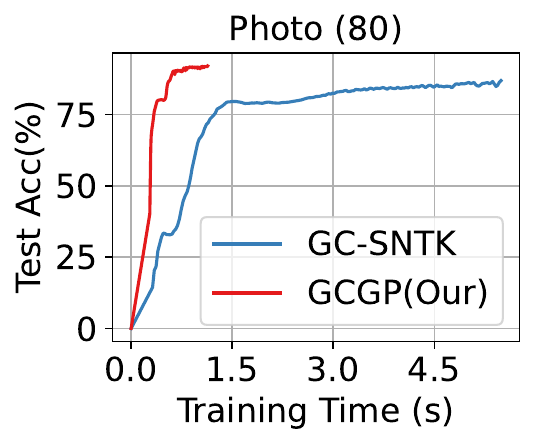}
  }
  \subfigure[Computers (100)]
  {
    \includegraphics[width=0.14\textwidth]{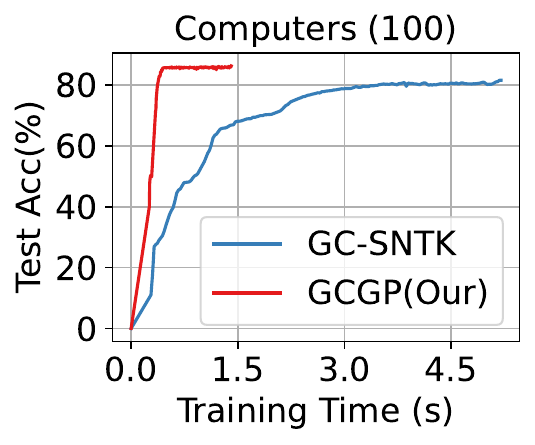}
  }
  \subfigure[\wl{Ogbn-arxiv (454)}]
  {
    \includegraphics[width=0.14\textwidth]{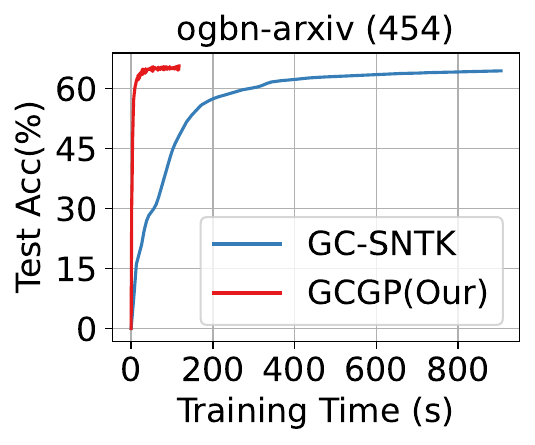}
  }
  
  \subfigure[Cora (140)]
  {
    \includegraphics[width=0.14\textwidth]{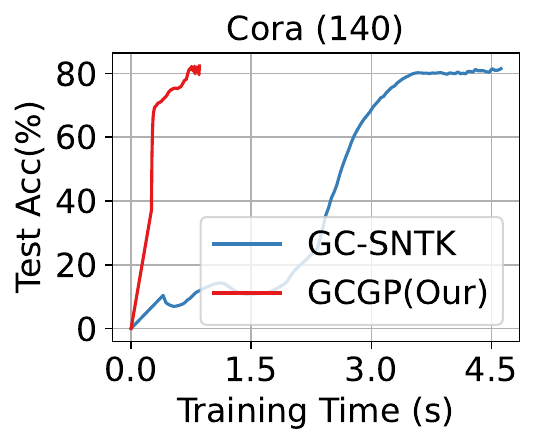}
  }
  \subfigure[Citeseer (120)]
  {
    \includegraphics[width=0.14\textwidth]{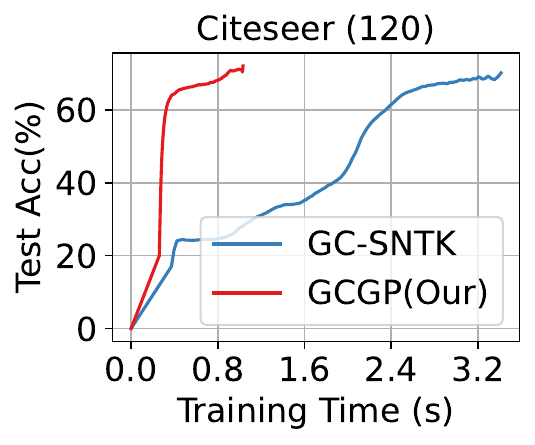}
  }
  \subfigure[Pubmed (60)]
  {
    \includegraphics[width=0.14\textwidth]{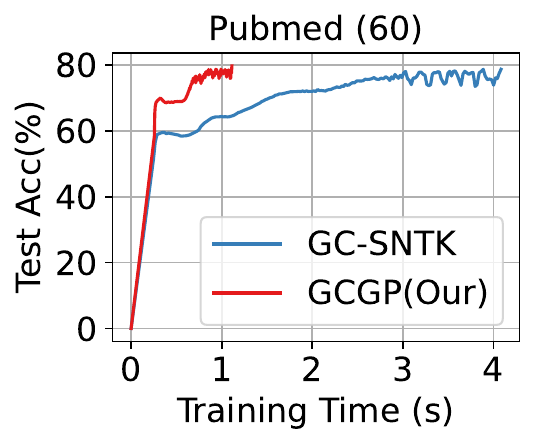}
  }
  \subfigure[Photo (160)]
  {
    \includegraphics[width=0.14\textwidth]{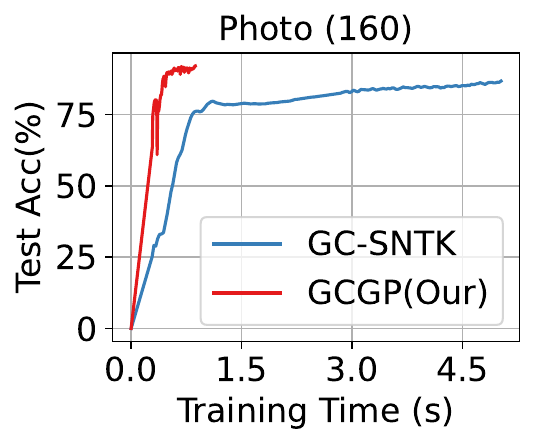}
  }
  \subfigure[Computers (100)]
  {
    \includegraphics[width=0.14\textwidth]{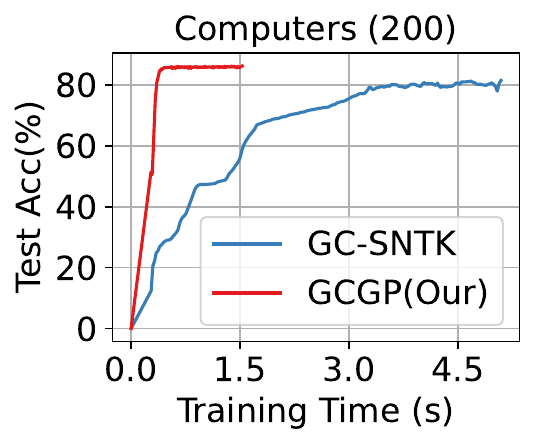}
  }
    \subfigure[\wl{Ogbn-arxiv (909)}]
  {
    \includegraphics[width=0.14\textwidth]{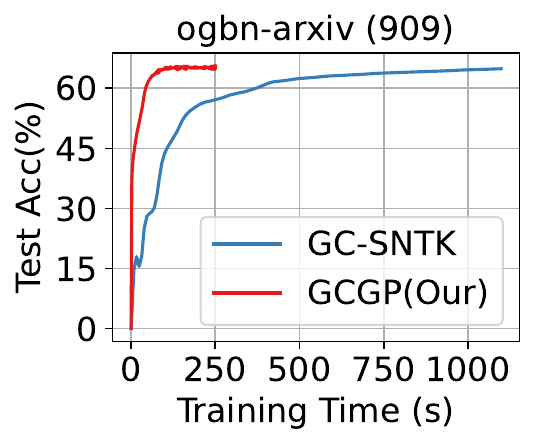}
  }
  \caption{\wl{Condensation efficiency comparison between the proposed GCGP method and GC-SNTK is conducted on five datasets. The x-axis represents training time, while the y-axis indicates test accuracy. The results show that GCGP consistently achieves faster training times than GC-SNTK across all condensation scales on these datasets.}}
  \label{fig:time_acc}
\end{figure*}



To further assess the efficiency of the proposed method, we compare GCGP with the kernel-based baseline GC-SNTK. The results, shown in Figure~\ref{fig:time_acc}, offer a comprehensive view of their performance across different datasets and condensation scales. Experiments are conducted on six datasets—Cora, Citeseer, Pubmed, Photo, Computers, and Ogbn-arxiv—under three condensation ratios, resulting in 18 settings. For each setting, we record the classification accuracy with respect to the training time required by the condensation process.

In Figure~\ref{fig:time_acc}, the $x$-axis indicates condensation time, while the $y$-axis shows the corresponding test accuracy. As observed, GCGP consistently achieves higher accuracy, demonstrating both faster condensation and better performance. Across most datasets, it outperforms GC-SNTK in terms of classification accuracy, highlighting the effectiveness of GCGP in improving both efficiency and condensation quality.

Additionally, Table~\ref{tab:times} summarizes the time consumed by the two methods across the 15 experimental settings, along with the speedup achieved by GCGP relative to GC-SNTK. The data clearly show that the proposed method is consistently more than three times faster than GC-SNTK while maintaining the quality of the condensed graphs. This stability and efficiency further underscore the robustness of the GCGP method.

In summary, the experimental results across all datasets confirm the significant efficiency improvements achieved by the proposed GCGP method. It consistently outperforms the fastest existing method, GC-SNTK, by more than three times in speed, thereby enhancing its scalability and broadening its potential application scenarios.
}

\begin{table}[]
\centering
\caption{\wl{Condensation time comparison of the GC-SNTK and the GCGP}}
\label{tab:times}
\renewcommand{\arraystretch}{1.2}
\begin{tabular}{ccccc}
\hline
Dataset                                                                          & Size         & GC-SNTK (s) & GCGP(s)  & Speedup \\ \hline
\multirow{3}{*}{Cora}                                                            & 1.30\% (35)  & 5.5843      & 0.7672   & 7.28×   \\
                                                                                 & 2.60\% (70)  & 5.0616      & 0.8383   & 6.04×   \\
                                                                                 & 5.20\% (140) & 4.6553      & 0.8595   & 5.42×   \\ \hline
\multirow{3}{*}{Citeseer}                                                        & 0.90\%(30)   & 5.8787      & 1.0806   & 5.44×   \\
                                                                                 & 1.80\%(60)   & 4.7093      & 0.7389   & 6.37×   \\
                                                                                 & 3.61\%(120)  & 3.4389      & 1.0386   & 3.31×   \\ \hline
\multirow{3}{*}{Pubmed}                                                          & 0.08\% (15)  & 4.1034      & 0.8610   & 4.77×   \\
                                                                                 & 0.15\% (30)  & 4.7844      & 1.0022   & 4.77×   \\
                                                                                 & 0.30\% (60)  & 4.1030      & 1.1172   & 3.67×   \\ \hline
\multirow{3}{*}{\begin{tabular}[c]{@{}c@{}}Amazon\\      Photo\end{tabular}}     & 0.5\% (40)   & 4.5685      & 1.1389   & 4.01×   \\
                                                                                 & 1.0\% (80)   & 5.5139      & 1.1430   & 4.82×   \\
                                                                                 & 2.1\% (160)  & 5.0429      & 0.8767   & 5.75×   \\ \hline
\multirow{3}{*}{\begin{tabular}[c]{@{}c@{}}Amazon\\      Computers\end{tabular}} & 0.4\% (50)   & 4.6563      & 0.5787   & 8.05×   \\
                                                                                 & 0.7\% (100)  & 5.2212      & 1.4125   & 3.71×   \\
                                                                                 & 1.5\% (200)  & 5.1126      & 1.5352   & 3.33×   \\ \hline
\multirow{3}{*}{Ogbn-arxiv}                                                      & 0.05\% (90)  & 842.0065    & 97.2541  & 8.66×   \\
                                                                                 & 0.25\% (454) & 904.8845    & 197.3606 & 4.58×   \\
                                                                                 & 0.5\% (909)  & 1098.2102   & 298.4082 & 3.68×   \\ \hline
\end{tabular}
\end{table}

\begin{figure}[ht]
  \centering
  \subfigure[Cora (X)]
  {
    \includegraphics[trim=70 30 30 45, clip,width=0.18\textwidth]{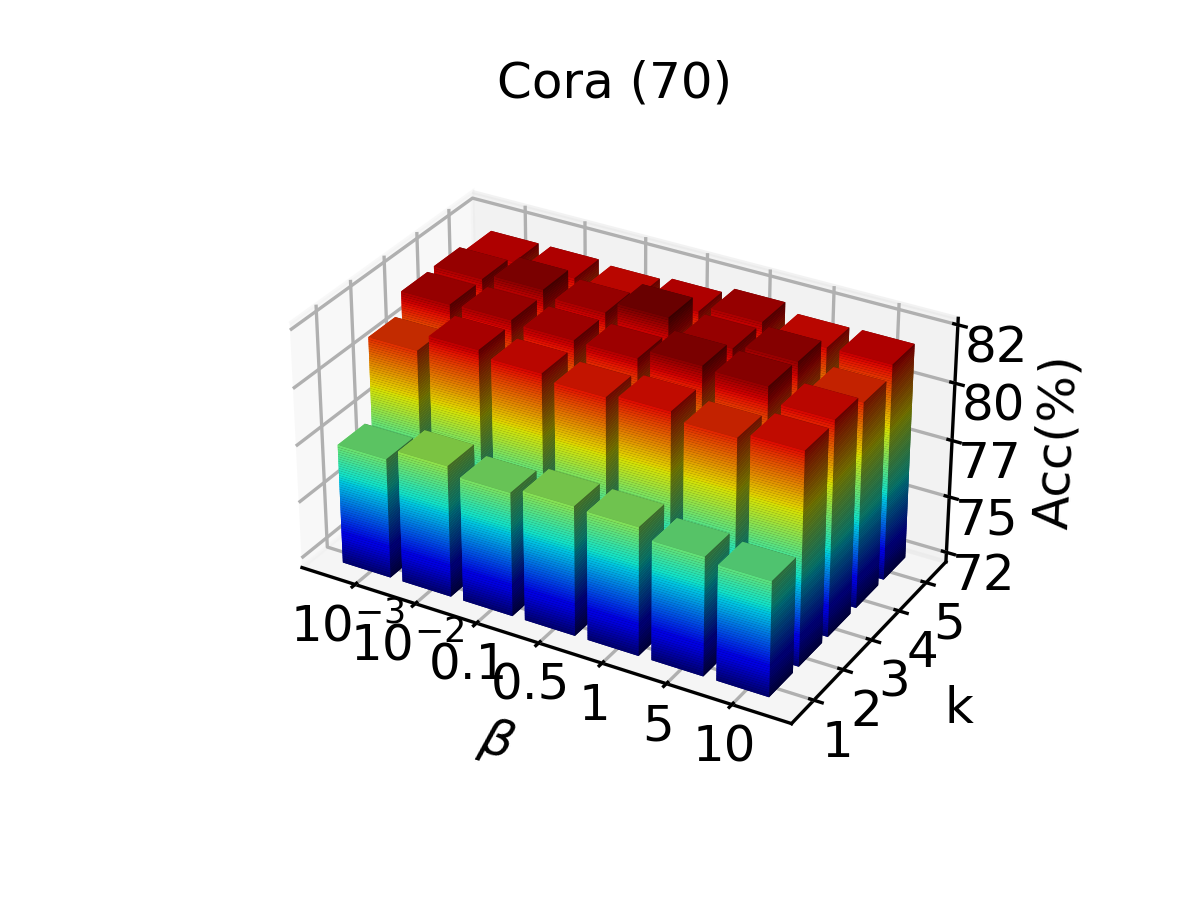}
  }
  \subfigure[Citeseer (X)]
  {
    \includegraphics[trim=70 30 30 45, clip,width=0.18\textwidth]{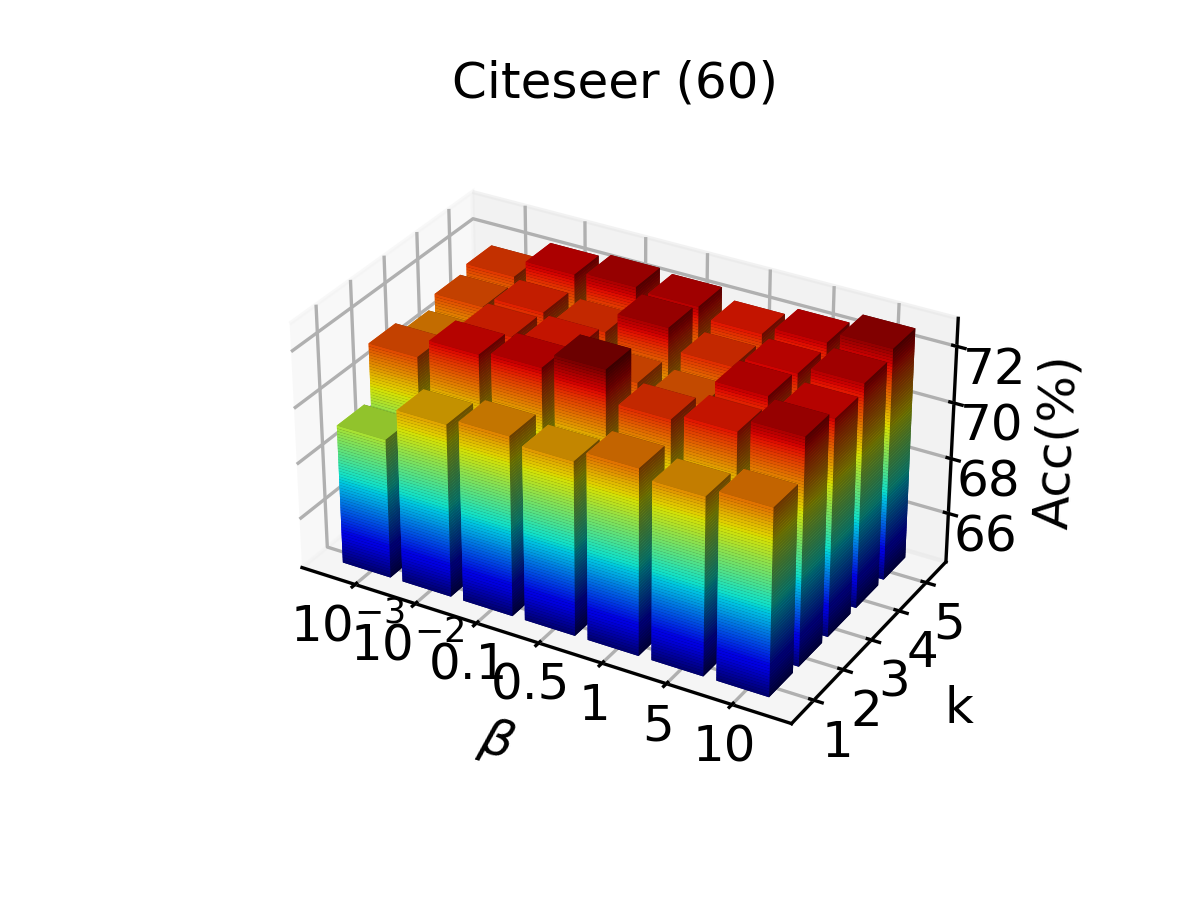}
  }
  \subfigure[Pubmed (X)]
  {
    \includegraphics[trim=70 30 30 45, clip,width=0.18\textwidth]{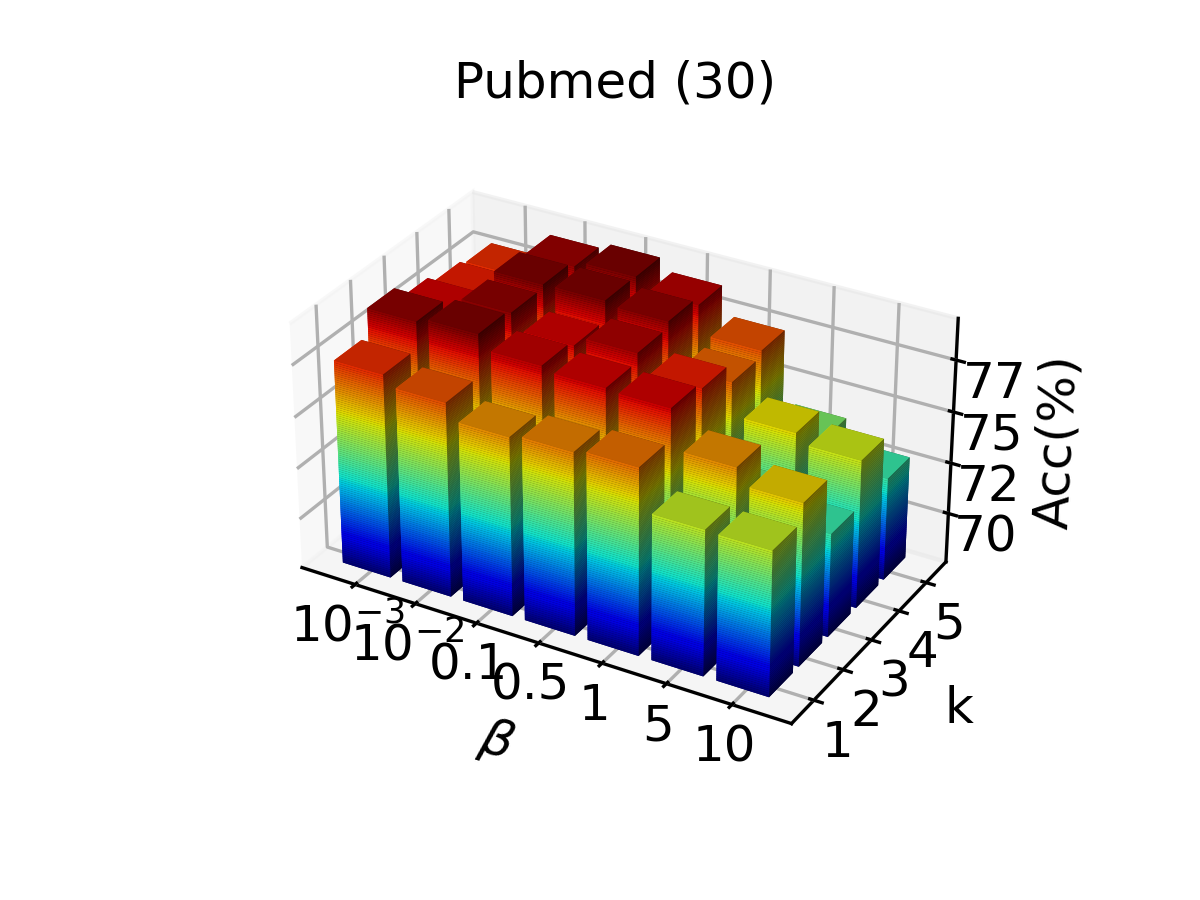}
  }
  \subfigure[Photo (X)]
  {
    \includegraphics[trim=70 30 30 45, clip,width=0.18\textwidth]{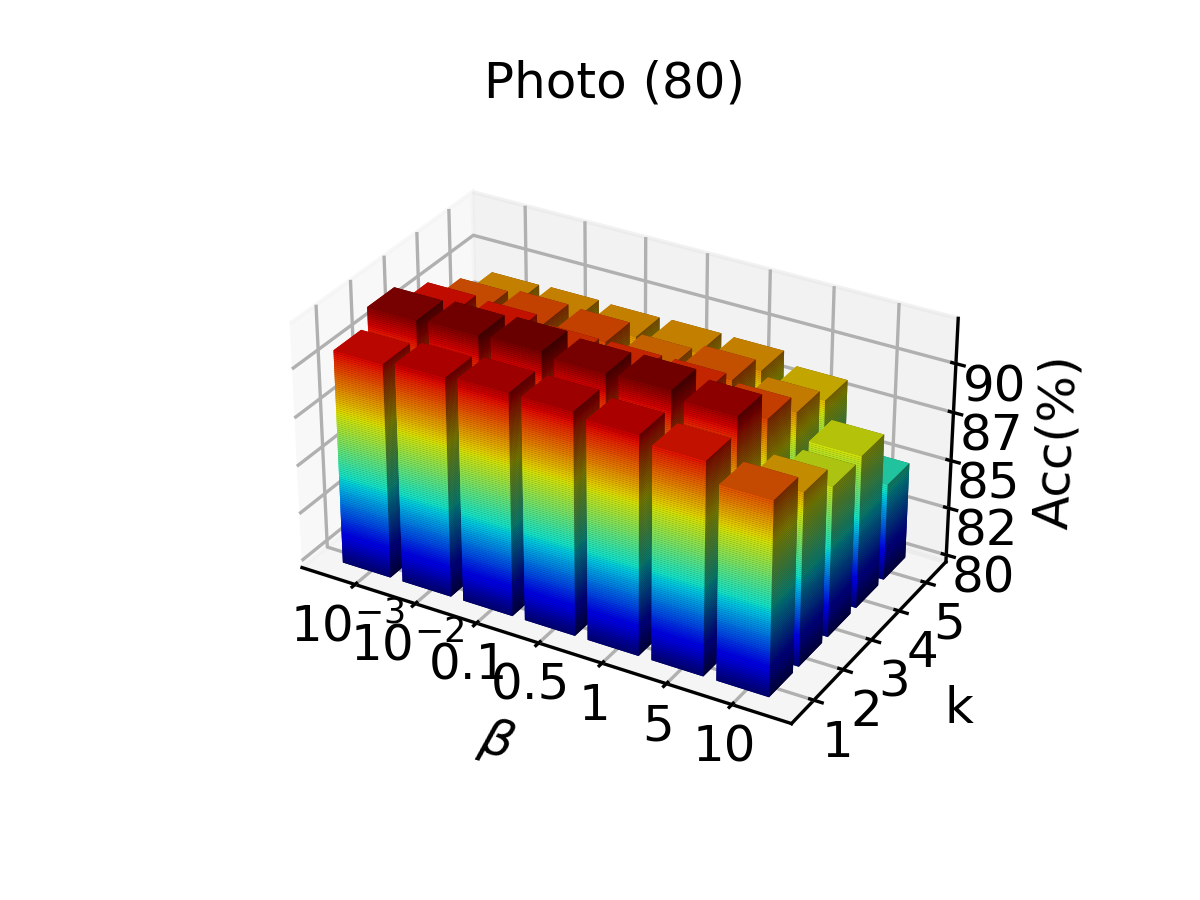}
  }
  \subfigure[Computers (X)]
  {
    \includegraphics[trim=70 30 30 45, clip,width=0.18\textwidth]{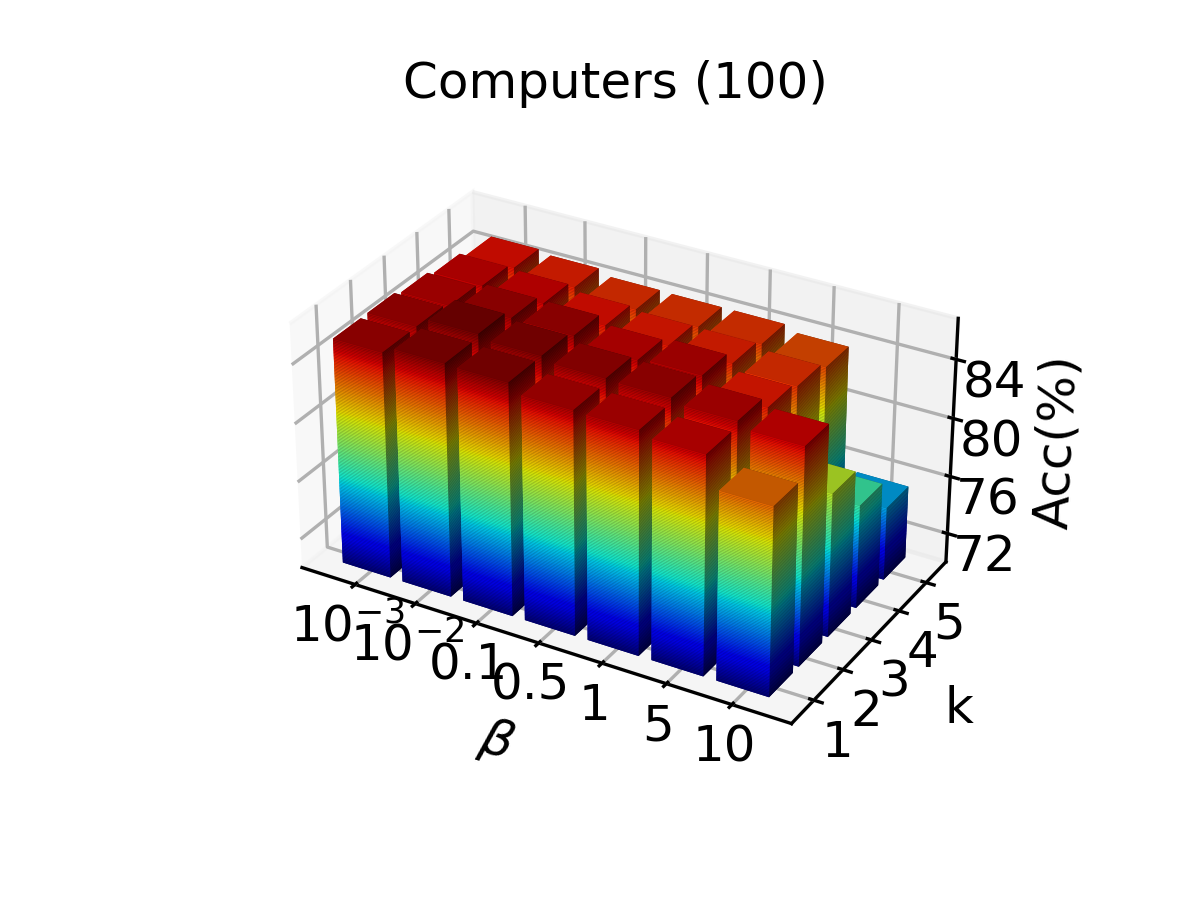}
  }
  \subfigure[Ogbn-arxiv (X)]
  {
    \includegraphics[trim=70 30 30 45, clip,width=0.18\textwidth]{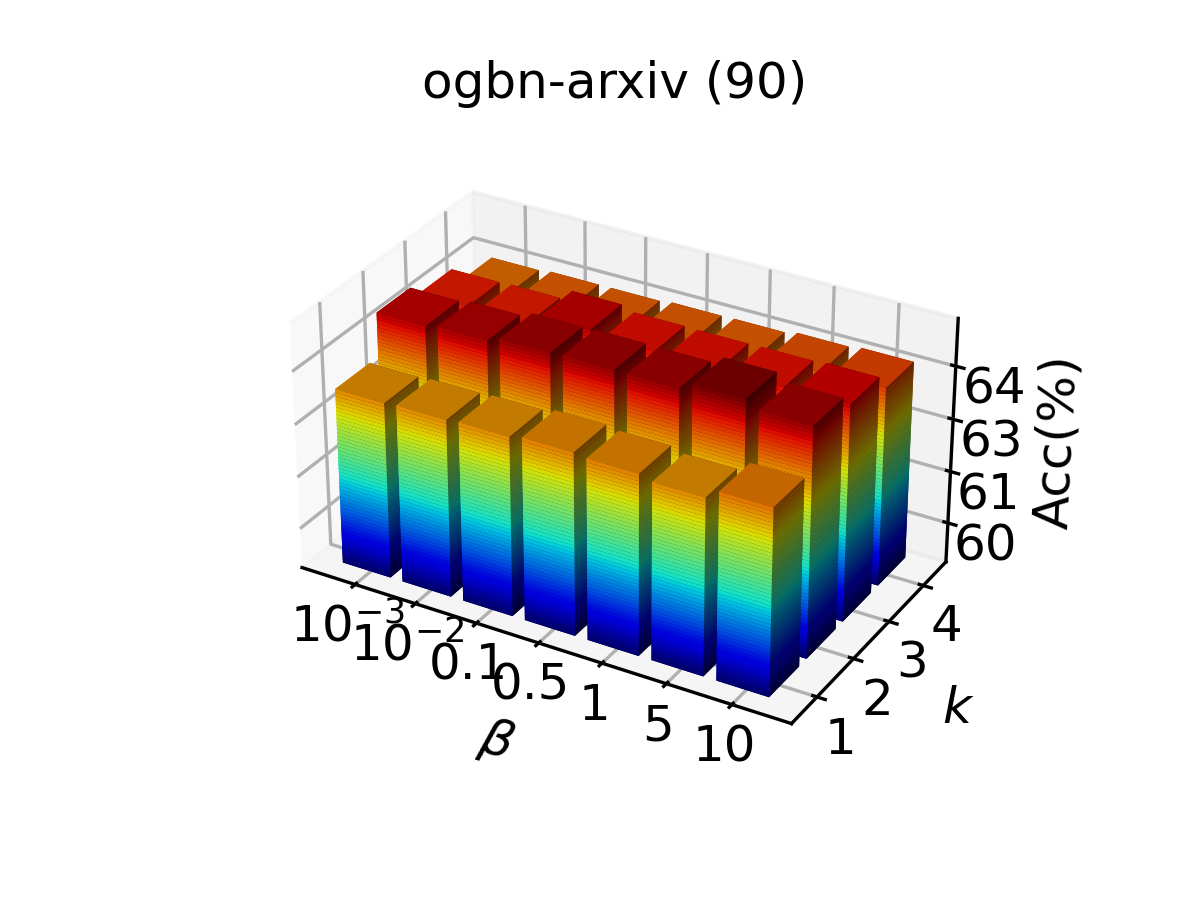}
  }
  \subfigure[Reddit (X)]
  {
    \includegraphics[trim=70 30 30 45, clip,width=0.18\textwidth]{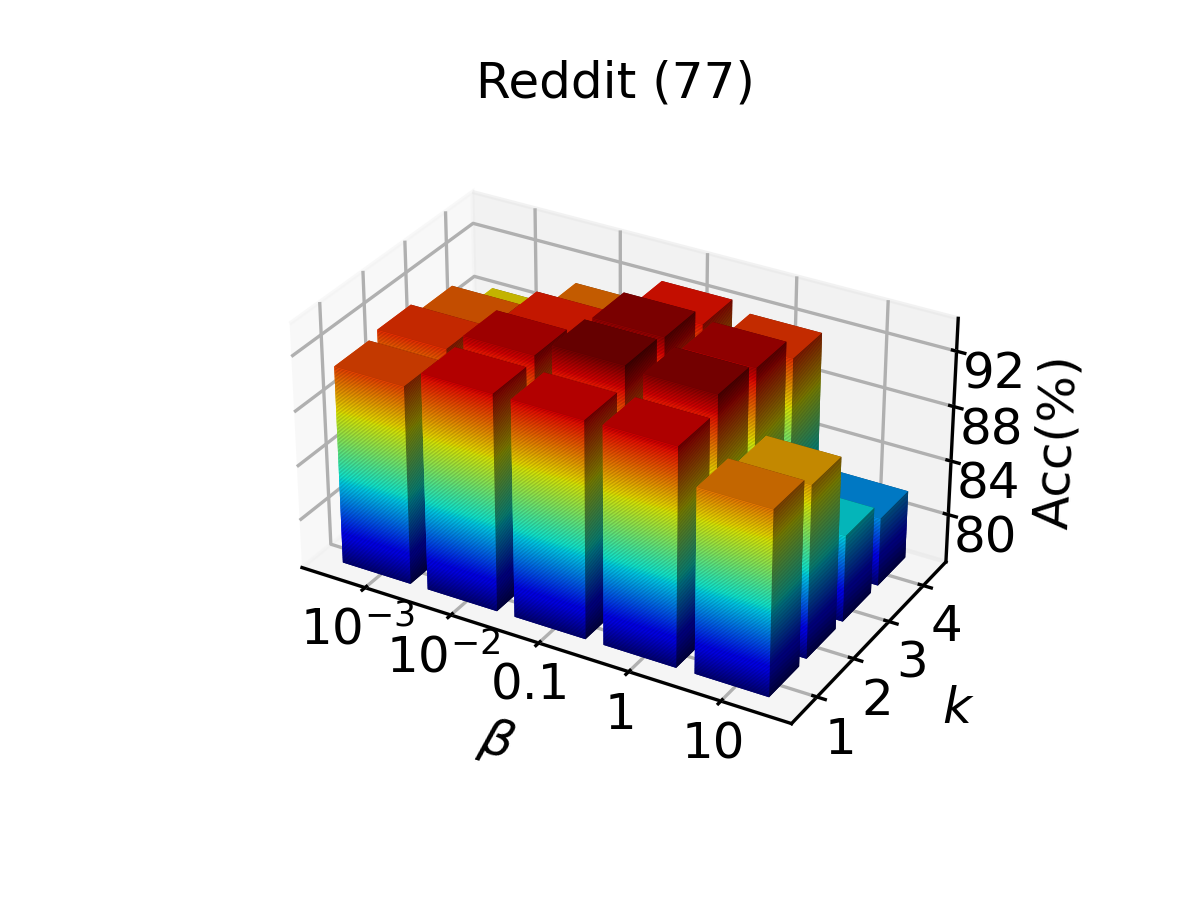}
  }
  \caption{Parameter sensitive analysis with condense only node features $X^{\mathcal{S}}$. The height of each bar represents the average classification accuracy on the condensed graph under the corresponding parameter settings. Darker bar colors indicate higher classification accuracy.}
  \label{fig:para1}
\end{figure}

\begin{figure}[ht]
  \centering

  \subfigure[Cora (X,A)]
  {
    \includegraphics[trim=70 30 30 45, clip,width=0.18\textwidth]{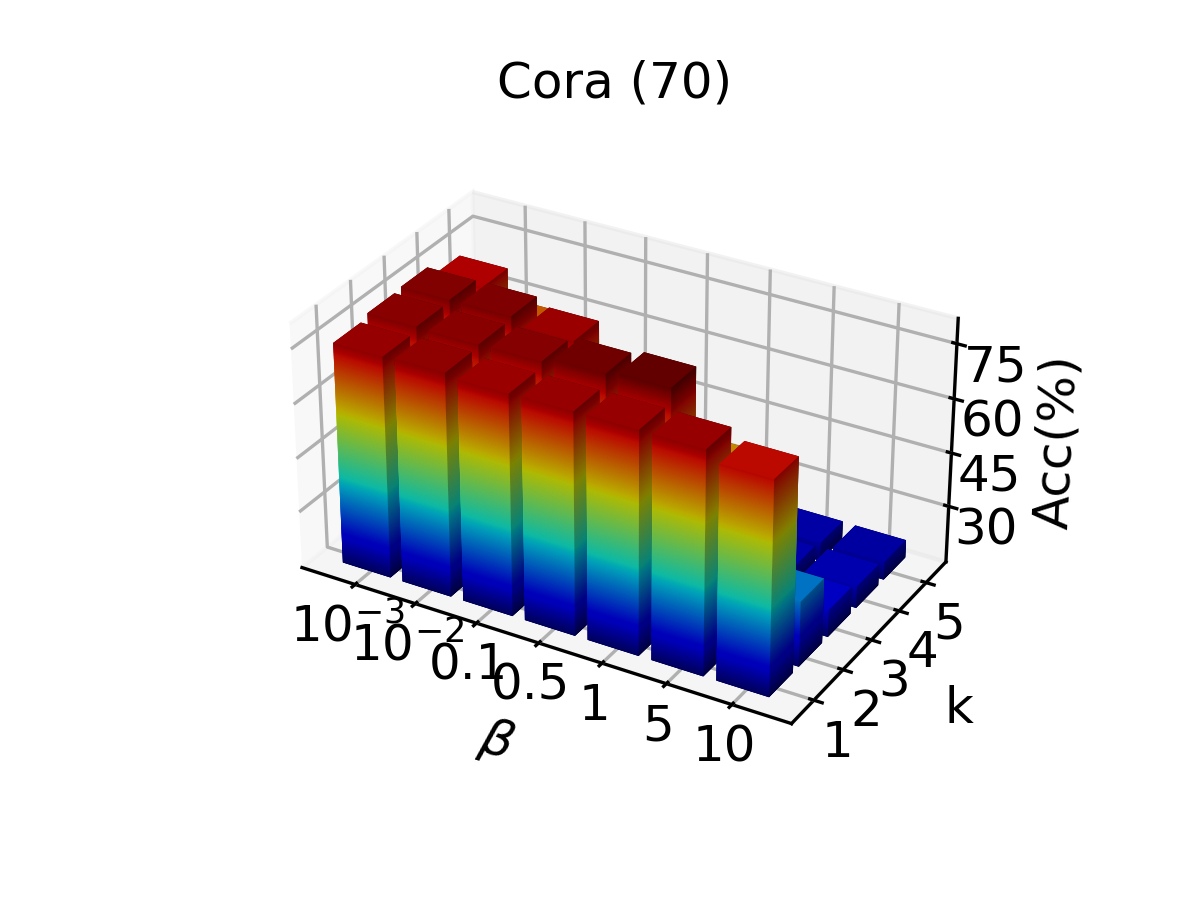}
  }
  \subfigure[Citeseer (X,A)]
  {
    \includegraphics[trim=70 30 30 45, clip,width=0.18\textwidth]{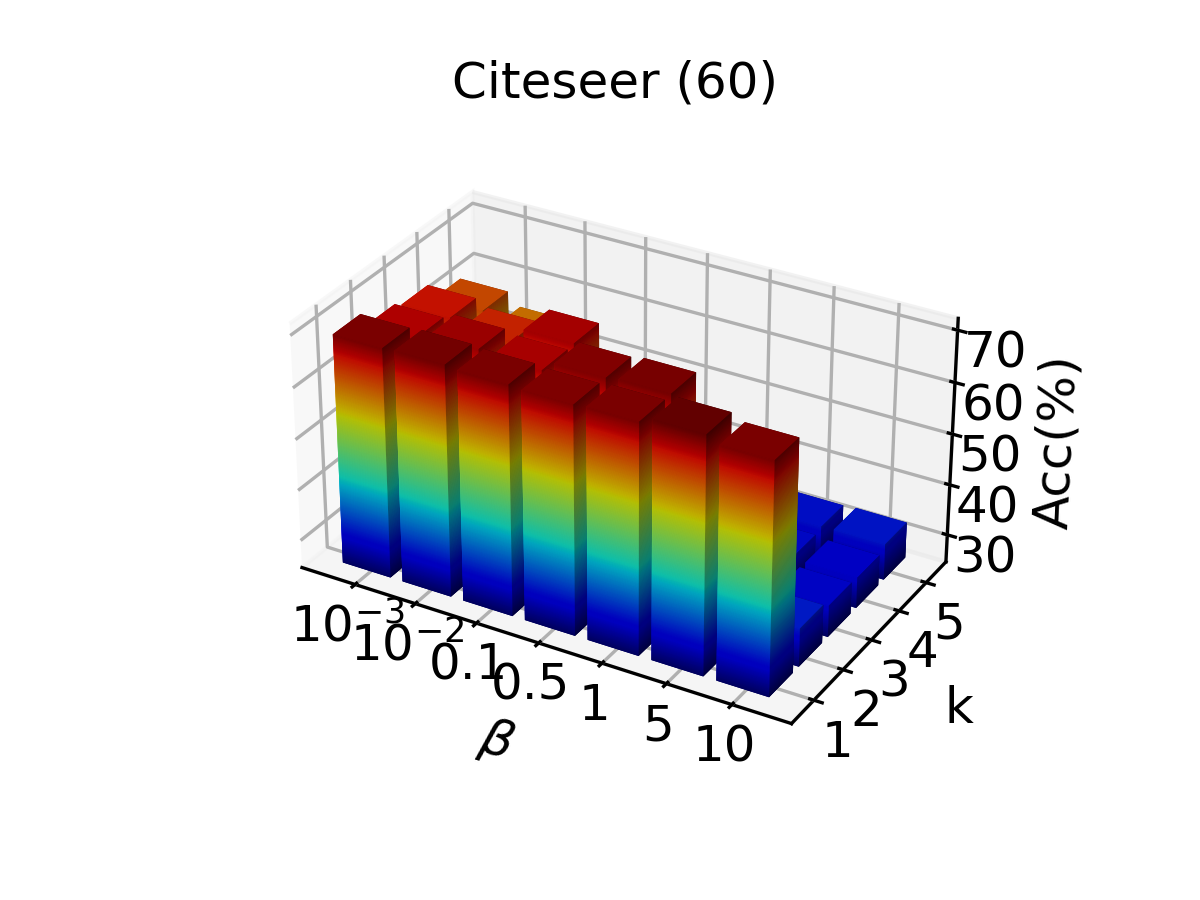}
  }
  \subfigure[Pubmed (X,A)]
  {
    \includegraphics[trim=70 30 30 45, clip,width=0.18\textwidth]{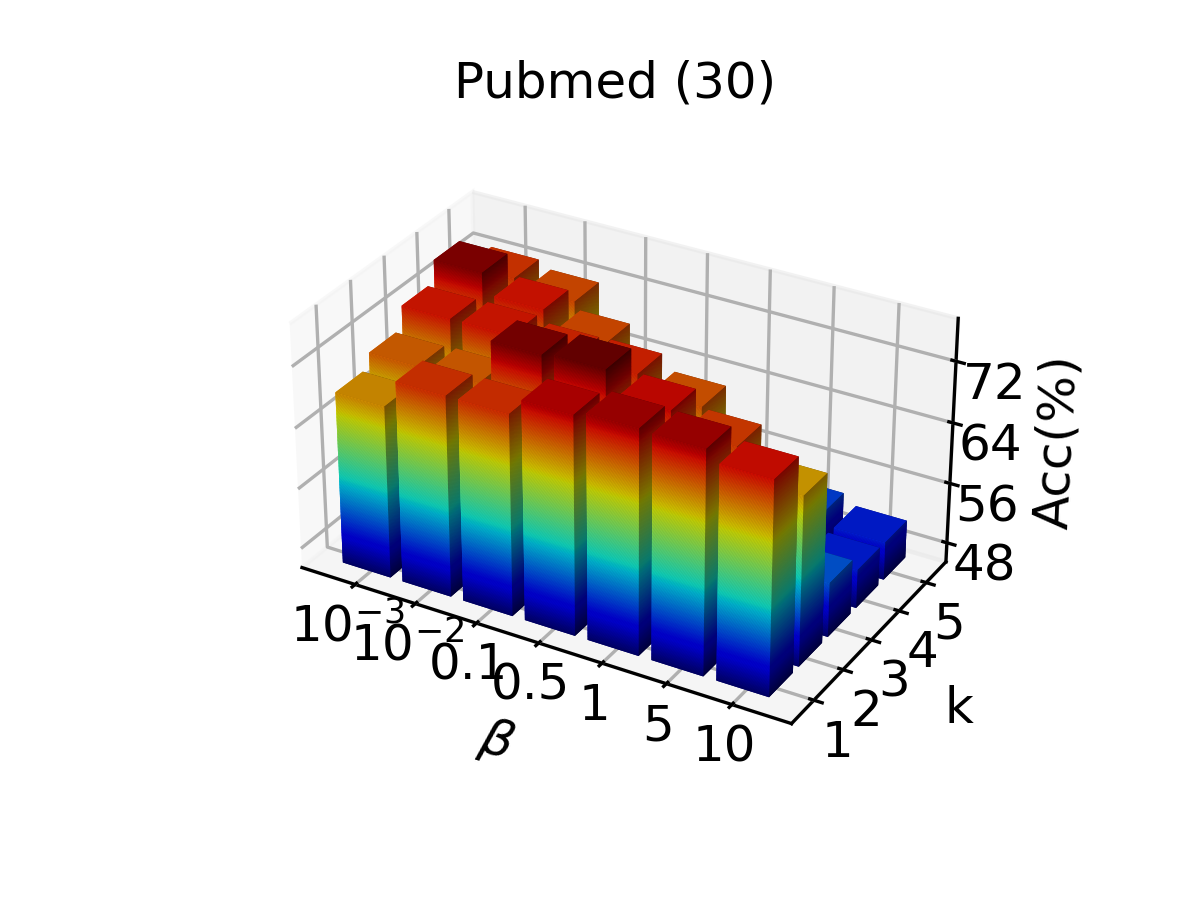}
  }
  \subfigure[Photo (X,A)]
  {
    \includegraphics[trim=70 30 30 45, clip,width=0.18\textwidth]{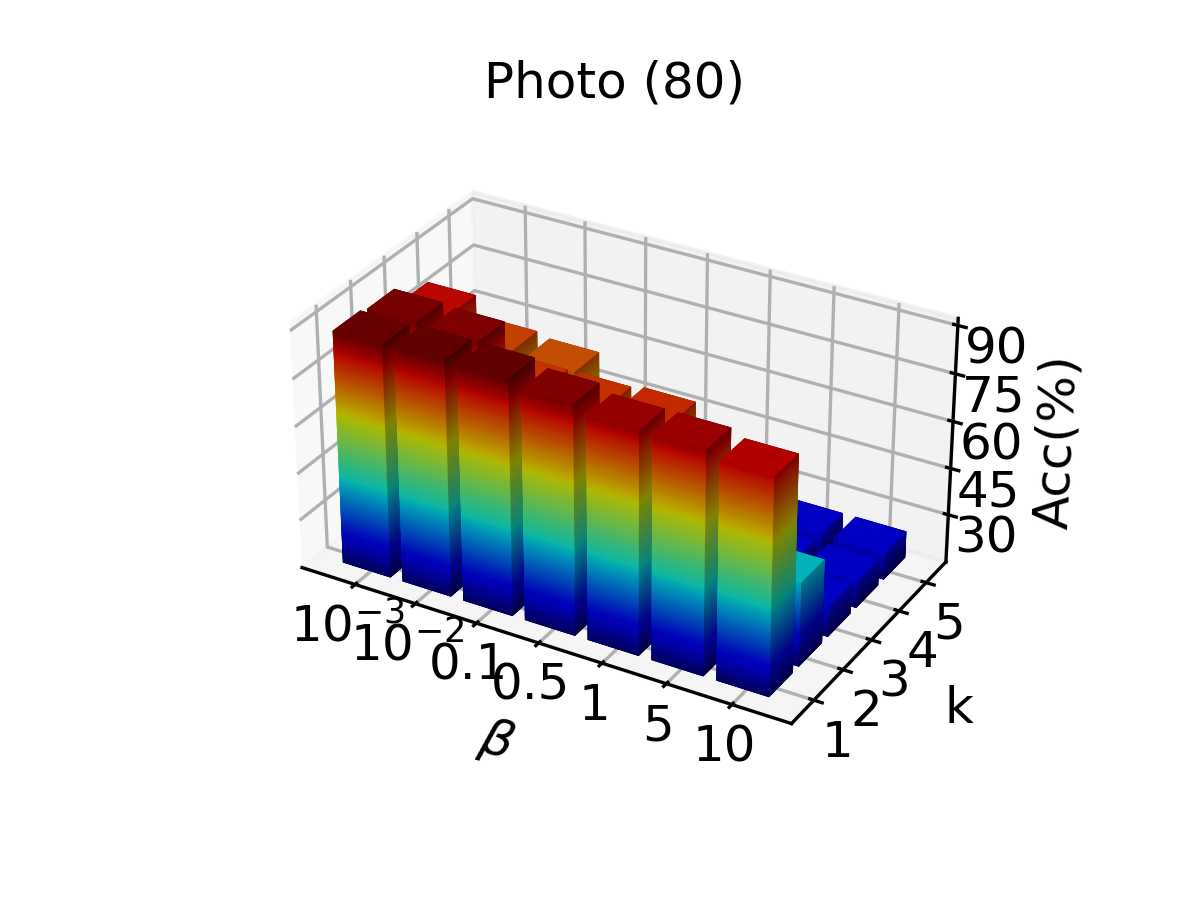}
  }
  \subfigure[Computers (X,A)]
  {
    \includegraphics[trim=70 30 30 45, clip,width=0.18\textwidth]{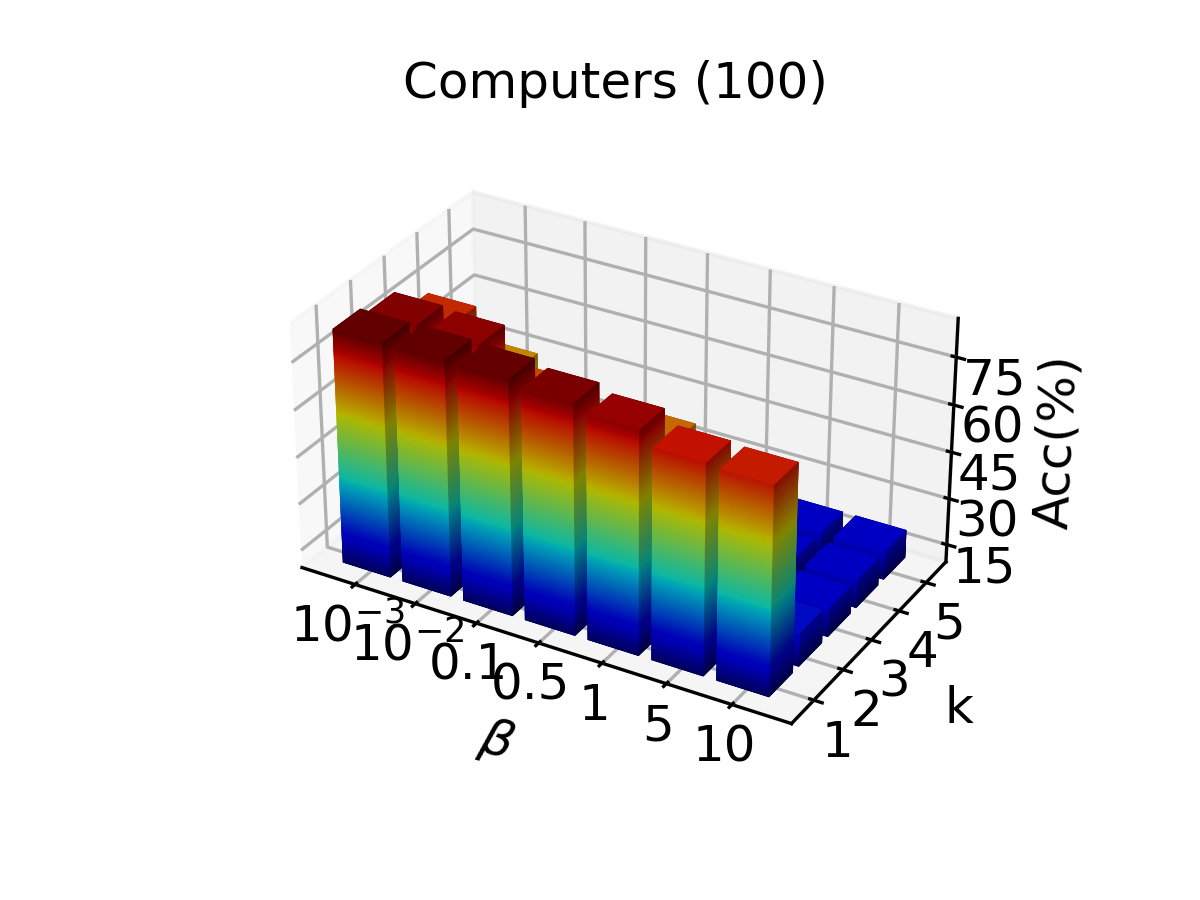}
  }
  \subfigure[Ogbn-arxiv (X,A)]
  {
    \includegraphics[trim=70 30 30 45, clip,width=0.18\textwidth]{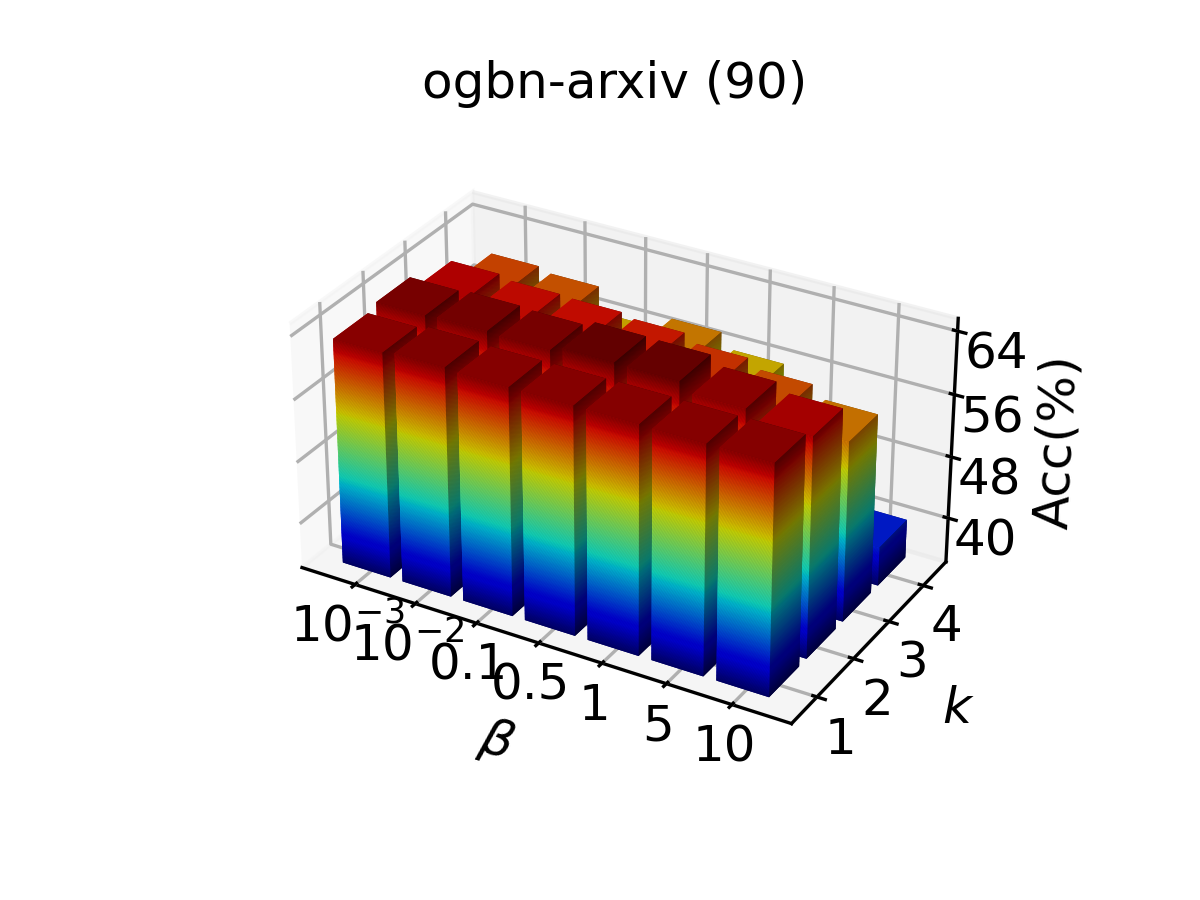}
  }
  \subfigure[Reddit (X,A)]
  {
    \includegraphics[trim=70 30 30 45, clip,width=0.18\textwidth]{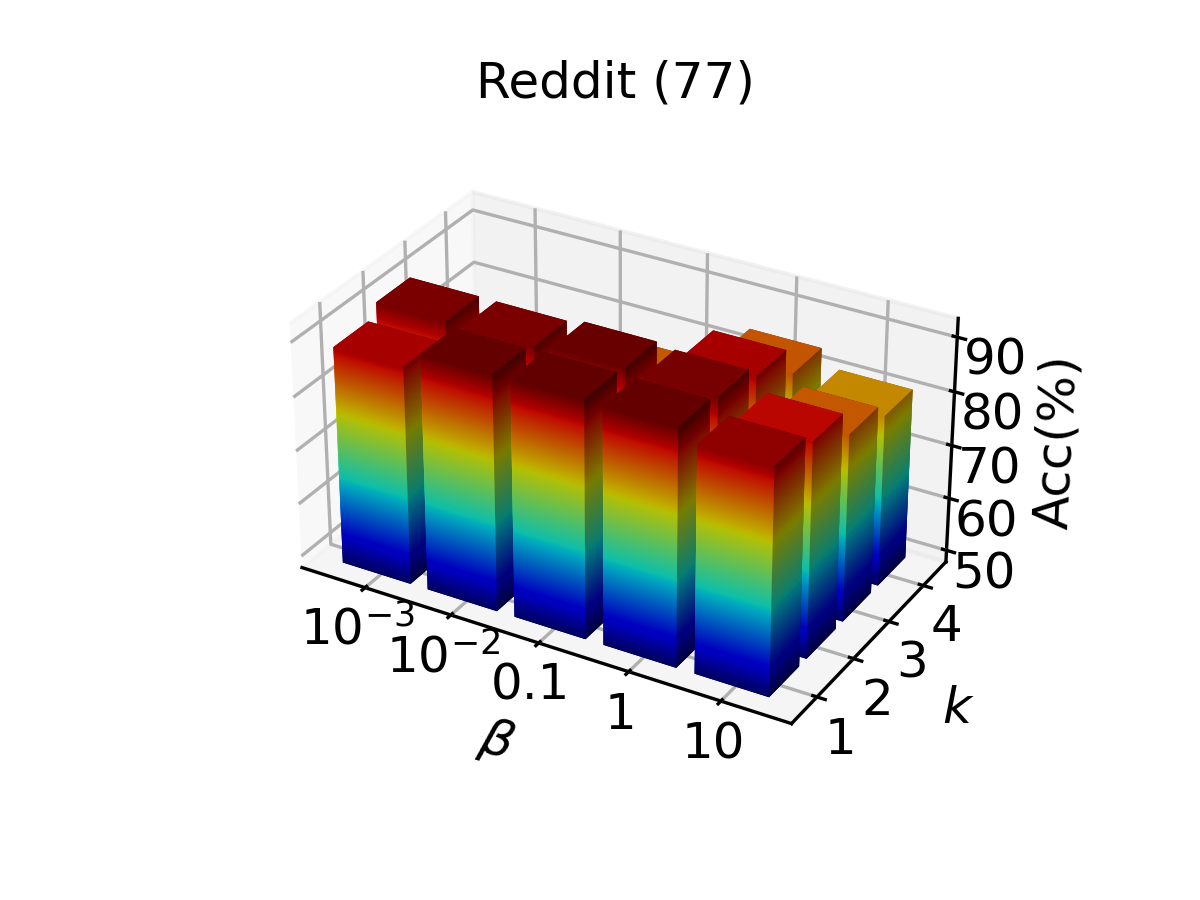}
  }
  \caption{Parameter sensitivity analysis condenses both node features $X^{\mathcal{S}}$ and structural information $A^{\mathcal{S}}$. 
  }
  \label{fig:para2}
\end{figure}


\subsection{Parameter Sensitivity Analysis}
To analyze the sensitivity of the proposed method to hyperparameters, we conduct experiments by varying two key parameters: $\beta$ in GP and the power $k$ of the adjacency matrix $A$ in the calculation of covariance function. These experiments are performed on multiple datasets, including Cora, Citeseer, Pubmed, Photo, Computer, and Ogbn-arxiv, with $\beta$ set to $\{ 10^{-3}, 10^{-2}, 10^{-1}, 0.5, 1, 5, 10 \}$. For the Reddit dataset, $\beta$ is adjusted to $\{ 10^{-3}, 10^{-2}, 10^{-1}, 1, 10 \}$. The parameter $k$ is tested with values $\{1, 2, 3, 4, 5\}$ for the Cora, Citeseer, Pubmed, Photo, and Computer datasets, and $\{1, 2, 3, 4\}$ for Ogbn-arxiv and Reddit. The condensation sizes for the datasets are fixed as follows: 70 for Cora, 60 for Citeseer, 30 for Pubmed, 80 for Photo, 100 for Computer, 90 for Ogbn-arxiv, and 77 for Reddit. All combinations of $\beta$ and $k$ are evaluated, and the average classification accuracy is computed over multiple iterations. The results, illustrated in Figure~\ref{fig:para1} and Figure~\ref{fig:para2}, show the classification accuracy achieved on the condensed graph for each parameter setting. In these figures, darker bar colors correspond to higher classification accuracy.

As shown in Figure~\ref{fig:para1}, which depicts the results for experiments condensing only node features, the accuracy is generally insensitive to parameter changes. For the Cora dataset, performance improves significantly and stabilizes when $k > 1$. The Citeseer dataset maintains high classification accuracy even for smaller $k$ values. The Pubmed dataset achieves favorable and stable results when $\beta$ is small, while the Photo dataset performs best and remains stable for smaller $k$ values. Among these datasets, the Computers dataset exhibits the most stable performance, with only minor variations when $\beta$ is large. Ogbn-arxiv is insensitive to both parameters, whereas Reddit performs poorly when $\beta$ and $k$ are large.

Figure~\ref{fig:para2} illustrates the results when both structural and feature information are condensed. A consistent trend can be observed: performance degrades significantly when both $k$ and $\beta$ are large. Conversely, the method achieves better results when both $k$ and $\beta$ are small.

In summary, the proposed method demonstrates varying degrees of sensitivity to hyperparameters across datasets. While some datasets exhibit stable performance regardless of parameter changes, others show performance differences depending on the parameter settings. These findings provide valuable insights into the selection of hyperparameters for different datasets.

\subsection{\wl{Generalization Across Different Models}}
\wl{To evaluate whether the graph data condensed by GCGP retains essential structural and semantic information for downstream tasks, we apply the condensed data to train a variety of GNN models. This setting allows us to assess the generalization ability of GCGP beyond the model used during condensation. The experimental results show that models trained on GCGP-condensed graphs consistently achieve competitive performance, demonstrating the robustness and transferability of the condensed data to different GNN architectures.}

Table~\ref{tab:gel} summarizes the generalization performance of graph data condensed by various methods—GCond, GC-SNTK, and the proposed GCGP—on multiple GNN models, including GCN~\cite{kipf2016semi}, SGC~\cite{wu2019simplifying}, APPNP~\cite{gasteiger2018predict}, GraphSAGE~\cite{hamilton2017inductive}, and KRR~\cite{vovk2013kernel}, across the Cora, Citeseer, and Pubmed datasets. Notably, the proposed GCGP method demonstrates competitive performance, achieving the highest or near-highest average accuracy across a majority of datasets and models. On the Cora dataset, GCGP achieves the highest average accuracy of 78.7\%, marginally surpassing the performance of GC-SNTK, which records an average accuracy of 78.0\%. It demonstrates strong generalization across architectures, achieving an accuracy of 82.6\% with KRR and 79.0\% with SGC. In contrast, baseline methods such as GCN-based condensation and APPNP-based condensation show significantly lower average accuracies of 30.1\% and 53.7\%, respectively. These results underscore the robustness of GCGP compared to alternative approaches.

On the Citeseer dataset, GCGP achieves the highest average accuracy of 68.4\%, outperforming GC-SNTK, which achieves 42.2\%, and matching the performance of SGC-based condensation, which records 68.2\%. Its performance is particularly strong with KRR, achieving an accuracy of 72.3\%, and with SGC, achieving 68.6\%. These results demonstrate the stable performance of GCGP across different models. In contrast, GC-SNTK exhibits limited effectiveness on this dataset, highlighting the adaptability of GCGP to varying graph structures.

On the Pubmed dataset, GCGP achieves an average accuracy of 71.4\%, which is competitive with GC-SNTK, achieving 75.7\%, and SGC-based condensation, achieving 76.2\%. The method performs strongly with APPNP and KRR, achieving accuracies of 77.0\% and 79.2\%, respectively. However, its accuracy with SGC is comparatively lower at 50.9\%. Despite this variation, GCGP demonstrates consistent performance across models, reflecting its reliability in diverse scenarios.

In conclusion, GCGP demonstrates robust generalization performance across all datasets and models. Its results suggest its effectiveness in preserving critical graph structures and features. Compared to baseline methods, GCGP consistently delivers superior outcomes, making it a reliable approach for graph data condensation in diverse downstream tasks.

{
\setlength{\tabcolsep}{3pt}
\begin{table}[]
\centering
\caption{\wl{Generalization of the condensed graph data. We evaluate the generalization of condensed graph data by training various GNNs on data derived from different methods. Model performance is assessed through test set classification accuracy.}}
\label{tab:gel}
\renewcommand{\arraystretch}{1.2}
\begin{tabular}{cclcccccc}
\hline
Dataset                                                                         & \multicolumn{2}{c}{Method}      & GCN  & SGC  & APPNP & SAGE & KRR  & Avg.                         \\ \hline
                                                                                &                         & GCN   & 31.9 & 34.8 & 33.5  & 32.6 & 17.5 & 30.1                         \\
                                                                                &                         & SGC   & 81.1 & 80.6 & 80.5  & 71.0 & 78.8 & \cellcolor[HTML]{D8E4BC}78.4 \\
                                                                                &                         & APPNP & 31.9 & 50.3 & 75.7  & 51.0 & 59.8 & 53.7                         \\
                                                                                & \multirow{-4}{*}{Gcond} & SAGE  & 63.6 & 41.2 & 68.9  & 53.5 & 26.5 & 50.7                         \\ \cline{2-3}
                                                                                & \multicolumn{2}{c}{GC-SNTK}     & 77.8 & 77.9 & 74.6  & 78.0 & 82.1 & \cellcolor[HTML]{EBF1DE}78.1 \\
\multirow{-6}{*}{\begin{tabular}[c]{@{}c@{}}Cora\\      (140)\end{tabular}}     & \multicolumn{2}{c}{GCGP}        & 79.7 & 79.0 & 74.1  & 78.1 & 82.6 & \cellcolor[HTML]{C4D79B}78.7 \\ \hline
                                                                                &                         & GCN   & 27.6 & 26.9 & 33.3  & 24.4 & 14.8 & 25.4                         \\
                                                                                &                         & SGC   & 72.6 & 65.4 & 71.5  & 62.2 & 69.5 & \cellcolor[HTML]{D8E4BC}68.2 \\
                                                                                &                         & APPNP & 52.6 & 56.8 & 71.2  & 45.1 & 54.1 & \cellcolor[HTML]{EBF1DE}56.0 \\
                                                                                & \multirow{-4}{*}{GCond} & SAGE  & 38.6 & 23.6 & 64.1  & 44.3 & 43.4 & 42.8                         \\ \cline{2-3}
                                                                                & \multicolumn{2}{c}{GC-SNTK}     & 20.9 & 22.1 & 52.2  & 48.4 & 67.3 & 42.2                         \\
\multirow{-6}{*}{\begin{tabular}[c]{@{}c@{}}Citeseer\\      (120)\end{tabular}} & \multicolumn{2}{c}{GCGP}        & 68.5 & 68.6 & 66.6  & 66.2 & 72.3 & \cellcolor[HTML]{C4D79B}68.4 \\ \hline
                                                                                &                         & GCN   & 43.7 & 62.6 & 61.6  & 70.3 & 66.1 & 60.9                         \\
                                                                                &                         & SGC   & 78.4 & 74.9 & 77.6  & 76.0 & 74.1 & \cellcolor[HTML]{C4D79B}76.2 \\
                                                                                &                         & APPNP & 67.7 & 57.8 & 77.2  & 64.1 & 59.9 & 65.3                         \\
                                                                                & \multirow{-4}{*}{GCond} & SAGE  & 71.2 & 58.8 & 74.2  & 69.0 & 52.5 & 65.1                         \\ \cline{2-3}
                                                                                & \multicolumn{2}{c}{GC-SNTK}     & 73.4 & 74.5 & 76.6  & 73.2 & 79.4 & \cellcolor[HTML]{D8E4BC}75.4 \\
\multirow{-6}{*}{\begin{tabular}[c]{@{}c@{}}Pubmed\\      (60)\end{tabular}}    & \multicolumn{2}{c}{GCGP}        & 75.2 & 50.9 & 77.0  & 74.7 & 79.2 & \cellcolor[HTML]{EBF1DE}71.4 \\ \hline
\end{tabular}
\end{table}

}

\subsection{Ablation Study}

\wl{To evaluate the effectiveness of the proposed covariance function in modeling node similarity, we conduct a comprehensive ablation study comparing GCGP with several alternative covariance functions, including Dot Product, Neural Tangent Kernel (NTK)~\cite{jacot2018neural}, Structure-based NTK (SNTK)~\cite{wang2024fast}, and LightGNTK~\cite{xu2023kernel}. The results are summarized in Table~\ref{tab:abl} and Table~\ref{tab:abl_time}, covering both classification accuracy and condensation time across seven datasets.

As shown in Table~\ref{tab:abl}, GCGP consistently outperforms all baseline covariance functions in terms of classification accuracy, particularly on datasets such as Cora, Citeseer, Photo, Computers, and Reddit. For example, on Photo, GCGP achieves 92.0\% accuracy, significantly surpassing SNTK (88.5\%) and LightGNTK (89.4\%). Similarly, on Reddit, GCGP reaches 93.9\%, far exceeding the next best method (77.9\% by SNTK).

In addition to accuracy, GCGP also excels in computational efficiency. As shown in Table~\ref{tab:abl_time}, GCGP dramatically reduces condensation time compared to SNTK and LightGNTK. Notably, on large-scale datasets like Ogbn-arxiv and Reddit, GCGP achieves over 8$\times$ and 6$\times$ speedups respectively compared to LightGNTK, and over 30$\times$ and 28$\times$ speedups compared to SNTK. This efficiency gain is crucial for scaling to large graphs.

In summary, GCGP demonstrates clear advantages in both performance and efficiency. Its covariance function not only captures graph structure and semantics more effectively but also ensures faster condensation, making it highly practical for real-world graph learning tasks.
}

\begin{table}[]
\centering
\caption{\wl{Impact of Covariance Function Ablation on Accuracy Performance.}}
\label{tab:abl}
\renewcommand{\arraystretch}{1.2}
\setlength{\tabcolsep}{2pt}
\begin{tabular}{cccccc}
\hline
\multirow{2}{*}{\textbf{Dataset}} & \multicolumn{5}{c}{\textbf{Acc ± Std.}}                                                 \\ \cline{2-6} 
                                  & \textbf{Dot Product} & \textbf{NTK} & \textbf{SNTK}     & \textbf{LightGNTK} & \textbf{GCGP}     \\ \hline
Cora (70)                         & 57.3±0.5             & 59.1±1.4     & 82.4±0.5          & 81.6±0.4           & \textbf{82.5±0.7} \\
Citeseer (60)                     & 60.9±1.2             & 62.7±0.6     & 67.0±0.3          & 71.1±1.1           & \textbf{72.8±0.6} \\
Pubmed (30)                       & 72.7±0.4             & 69.9±1.2     & \textbf{79.3±0.3} & 78.8±0.5           & 79.2±0.7          \\
Photo (80)                        & 87.6±1.9             & 82.9±0.3     & 88.5±0.2          & 89.4±0.8           & \textbf{92.0±0.1} \\
Computers (100)                   & 80.7±1.2             & 72.5±0.9     & 83.1±0.6          & 83.5±0.5           & \textbf{86.4±0.3} \\
Ogbn-arxiv (90)                   & 50.6±0.5             & 51.3±0.1     & 64.2±0.2          & 60.5±0.4           & \textbf{65.7±0.3} \\
Reddit (77)                       & 66.3±0.1             & 66.3±0.2     & 77.9±0.9          & 71.8±1.5           & \textbf{93.9±0.1} \\ \hline
\end{tabular}
\end{table}

\begin{table}[]
\caption{\wl{The Time Comparison of Different Covariance Functions.}}
\label{tab:abl_time}
\renewcommand{\arraystretch}{1.2}
\centering
\begin{tabular}{cccc}
\hline
\multirow{2}{*}{\textbf{Dataset}} & \multicolumn{3}{c}{\textbf{Time (s)}}     \\ \cline{2-4} 
                                  & \textbf{SNTK} & \textbf{LightGNTK} & \textbf{GCGP}   \\ \hline
Cora (70)                         & 5.06          & 1.31               & \textbf{0.84}   \\
Citeseer (60)                     & 4.71          & 4.56               & \textbf{0.74}   \\
Pubmed (30)                       & 4.78          & 2.45               & \textbf{1.00}   \\
Photo (80)                        & 5.51          & 1.52               & \textbf{1.14}   \\
Computers (100)                   & 5.22          & 1.87               & \textbf{1.41}   \\
Ogbn-arxiv (90)                   & 871.52        & 772.39             & \textbf{97.25}  \\
Reddit (77)                       & 3844.36       & 854.96             & \textbf{135.68} \\ \hline
\end{tabular}
\end{table}

\section{Conclusion}


In this paper, we propose a novel GP-based framework for graph data condensation, addressing the computational challenges of existing bi-level methods. By treating the condensed graph as GP observations, our approach eliminates iterative GNN training, enhancing computational efficiency while maintaining predictive performance. To capture graph structure, we design a covariance function that incorporates $k$-hop message passing and represents local structures with minimal computational overhead. We propose using the binary concrete relaxation to optimize the adjacency matrix. This approach relaxes the discrete adjacency matrix into a differentiable one, enabling the efficient optimization of the graph structure. In summary, our contributions are: (1) integrating GP into graph condensation to eliminate iterative GNN training; (2) designing a covariance function to capture node similarities; (3) introducing the binary concrete relaxation for differentiable optimization of discrete graph structures; and (4) empirically validating the method on diverse datasets and baselines. Together, these contributions enhance the efficiency and practicality of graph condensation.



\bibliographystyle{ieeetr}
\bibliography{references}

\begin{IEEEbiography}[{\includegraphics[width=1in,height=1.25in,clip,keepaspectratio]{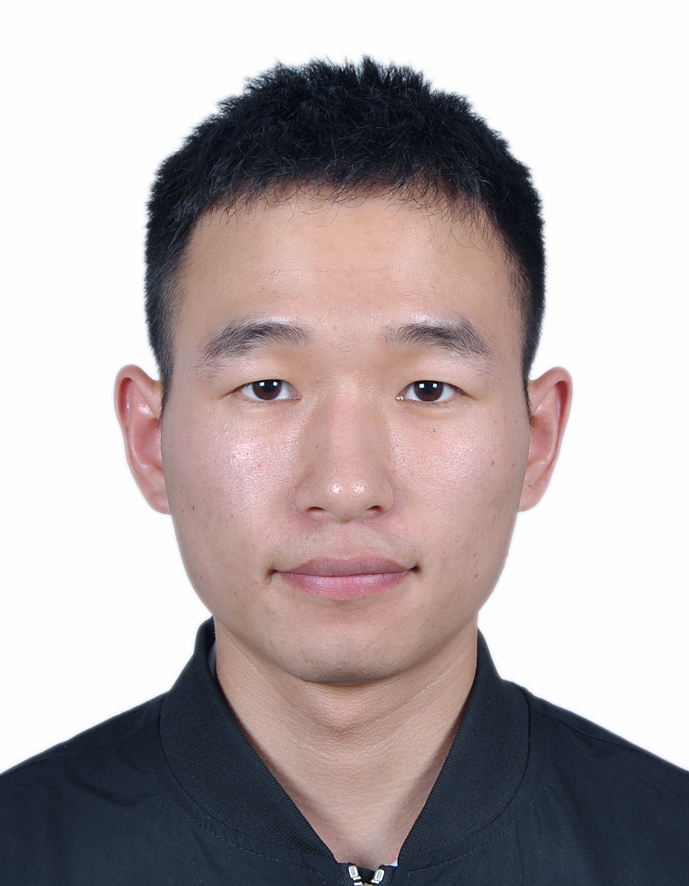}}]{Lin Wang} received the B.Eng. degree in aeronautical power engineering in 2019 and the M.Eng. degree in navigation, guidance, and control in 2022, both from Northwestern Polytechnical University, Xi'an, China. He is currently pursuing the Ph.D. degree in computer science with the Department of Computing, The Hong Kong Polytechnic University, Hong Kong SAR, China, under the supervision of Prof. Qing Li. His research interests include graph learning, recommender systems, LLM-based recommendation, and machine learning. He has published papers in venues such as WWW, TNNLS, and CIKM.
\end{IEEEbiography}

\begin{IEEEbiography}[{\includegraphics[width=1in,height=1.25in,clip,keepaspectratio]{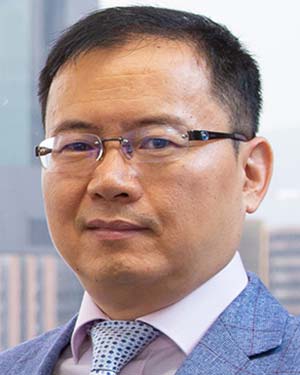}}]{Qing Li} (Fellow, IEEE) received the BEng degree in computer science from Hunan University, Changsha, China, and the MSc and PhD degrees from the University of Southern California, Los Angeles. He is currently a chair professor (Data Science) and the Head of the Department of Computing, the Hong Kong Polytechnic University. He is a fellow of IET,
a member of ACM SIGMOD and IEEE Technical Committee on Data Engineering. His research interests include object modeling, multimedia databases, social media, and recommender systems. He has been actively involved in the research community by serving as an associate editor and reviewer for technical journals, and as an organizer/co-organizer of numerous international conferences. He is the chairperson of the Hong Kong Web Society, and also served/is serving as an executive committee (EXCO) member of IEEE-Hong Kong Computer Chapter and ACM Hong Kong Chapter. In addition, he serves as a councilor of the Database Society of Chinese Computer Federation (CCF), a member of the Big Data Expert Committee of CCF, and is a Steering Committee member of DASFAA, ER, ICWL, UMEDIA, and WISE Society.
\end{IEEEbiography}

\end{document}